\documentclass[10pt,twocolumn,letterpaper]{article}

\usepackage[pagenumbers]{cvpr} 

\usepackage[dvipsnames]{xcolor}

\usepackage{float}
\usepackage{amssymb}
\usepackage{mathtools}
\usepackage{tikz}
\usepackage{pgfplots}

\usepackage{tabularray}
\usepackage{float}
\usepackage{makecell}
\usepackage{graphicx}
\usepackage{rotating}
\usepackage{amsmath,amssymb}
\usepackage{mdframed}
\usepackage{xcolor}
\usepackage{rotating}
\usepackage{tabularray}
\usepackage{graphbox} 
\usepackage{graphicx}
\usepackage{multirow}
\usepackage{rotating}
\usepackage{amsmath}
\usepackage{algorithm}
\usepackage{algpseudocode}
\usepackage{makecell}
\usepackage{pgfplots}
\usepackage{tikz}
\usepackage{float}
\usepackage{afterpage}
\usepackage{enumitem}
\usepackage{listings}
\usepackage{xcolor}
\usepackage{graphicx}
\usepackage{booktabs}

\DeclareMathOperator*{\argmin}{arg\,min}
\definecolor{cvprblue}{rgb}{0.21,0.49,0.74}
\usepackage[pagebackref,breaklinks,colorlinks,citecolor=cvprblue]{hyperref}

\def\paperID{*****} 
\def\confName{CVPR}
\def\confYear{2024}

\title{Robust Concept Erasure Using Task Vectors} 

\author{Minh Pham, Kelly O. Marshall, Chinmay Hegde, and Niv Cohen \\
New York Univeristy\\
{\tt\small \{mp5847,km3888,chinmay.h,nc3468\}@nyu.edu}
}

\begin{document}
\maketitle
\begin{abstract}
With the rapid growth of text-to-image models, a variety of techniques have been suggested to prevent undesirable image generations. Yet, these methods often only protect against specific user prompts and have been shown to allow unsafe generations with other inputs. Here we focus on \textit{unconditionally} erasing a concept from a text-to-image model rather than conditioning the erasure on the user's prompt. We first show that compared to input-dependent erasure methods, concept erasure that uses Task Vectors (TV) is more robust to unexpected user inputs, not seen during training. However, TV-based erasure can also affect the core performance of the edited model, particularly when the required edit strength is unknown. To this end, we propose a method called \textit{Diverse Inversion}, which we use to estimate the required strength of the TV edit. Diverse Inversion finds within the model input space a large set of word embeddings, each of which induces the generation of the target concept. We find that encouraging diversity in the set makes our estimation more robust to unexpected prompts. Finally, we show that Diverse Inversion enables us to apply a TV edit only to a subset of the model weights, enhancing the erasure capabilities while better maintaining the core functionality of the model\footnote{Our code can be found at \url{https://github.com/mnpham0417/prompt-agnostic-concept-erasure}}.
\end{abstract}    
\begin{figure*}[ht]
\centering
\fontsize{8}{1}\selectfont
\resizebox{0.8\textwidth}{!}{
\begin{tblr}{
  width = \linewidth,
  colspec = {Q[20]Q[80]Q[80]Q[80]Q[80]Q[80]Q[80]Q[80]Q[80]Q[80]Q[80]Q[80]}, 
  column{even} = {c},
  column{odd} = {c},
  rowsep=1pt,
  colsep=1pt,
  vline{12}={1-4}{},
}
& [-1.0,-0.8] & [-0.8,-0.6] & [-0.6,-0.4] & [-0.4,-0.2] & [-0.2,0.0] & [0.0,0.2] & [0.2,0.4] & [0.4,0.6] & [0.6,0.8] & [0.95, 0.98] & ``Van Gogh"    \\
\begin{sideways}\hspace{6pt}\makecell{ESD}\end{sideways} & \includegraphics[width=\linewidth,height=\linewidth]{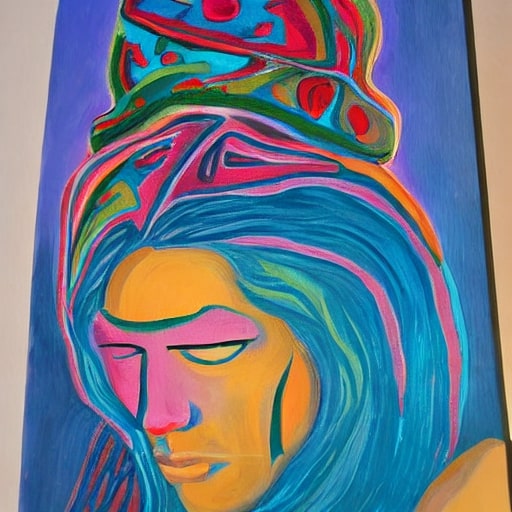} & \includegraphics[width=\linewidth,height=\linewidth]{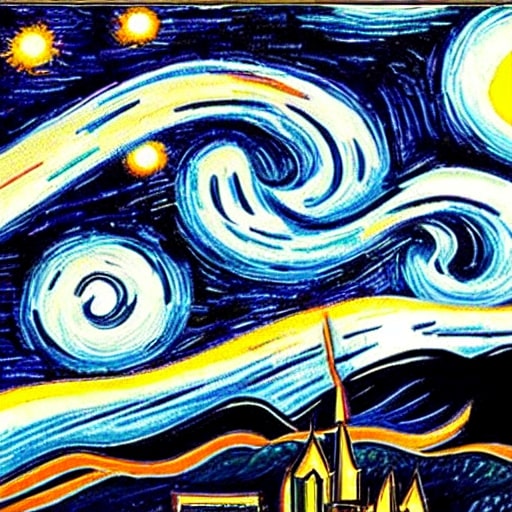} & \includegraphics[width=\linewidth,height=\linewidth]{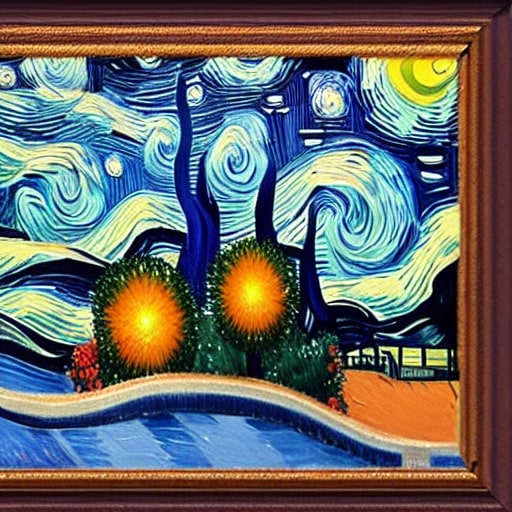} & \includegraphics[width=\linewidth,height=\linewidth]{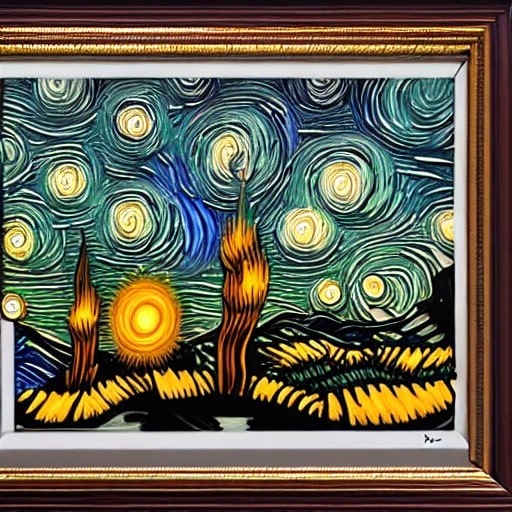} & \includegraphics[width=\linewidth,height=\linewidth]{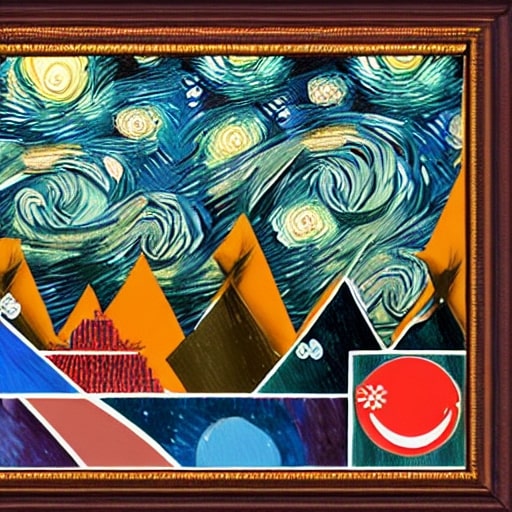}  & 
\includegraphics[width=\linewidth,height=\linewidth]{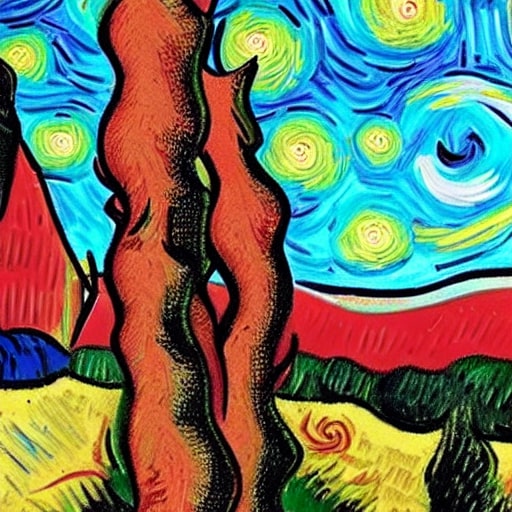} & \includegraphics[width=\linewidth,height=\linewidth]{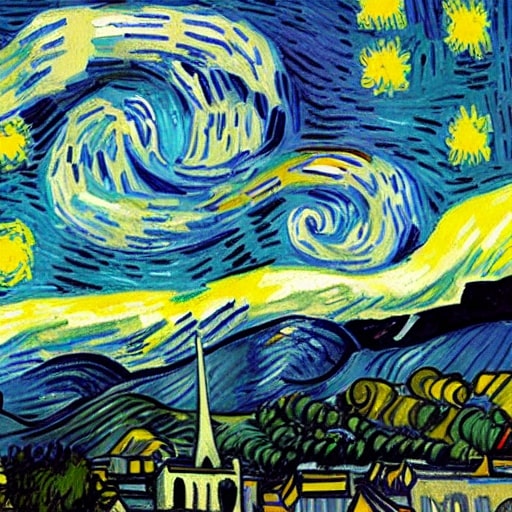} & \includegraphics[width=\linewidth,height=\linewidth]{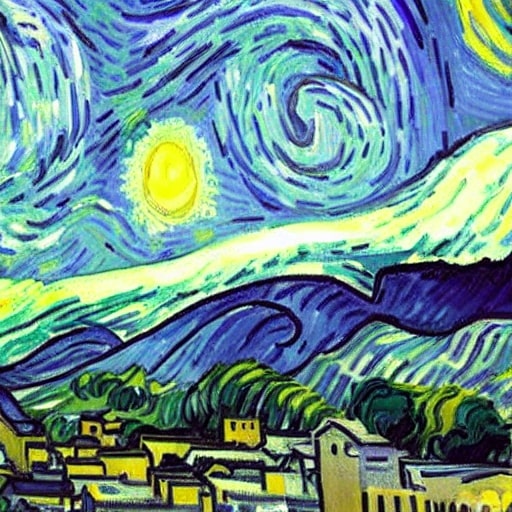} & \includegraphics[width=\linewidth,height=\linewidth]{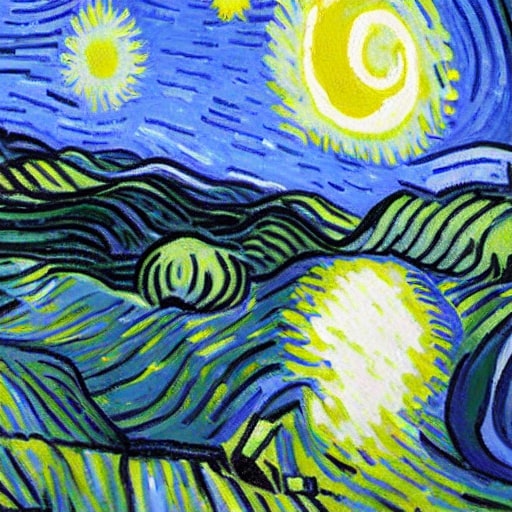} & \includegraphics[width=\linewidth,height=\linewidth]{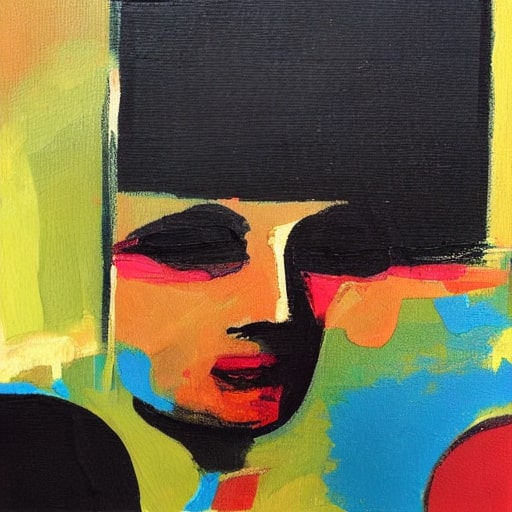} & \includegraphics[width=\linewidth,height=\linewidth]{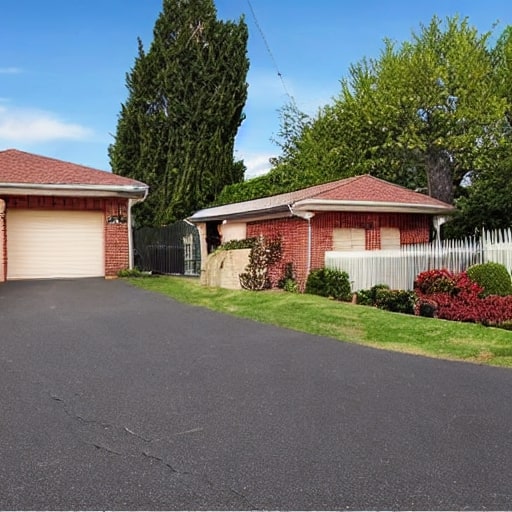} \\
\begin{sideways}\hspace{6pt}\makecell{SD 1.4}\end{sideways} & \includegraphics[width=\linewidth,height=\linewidth]{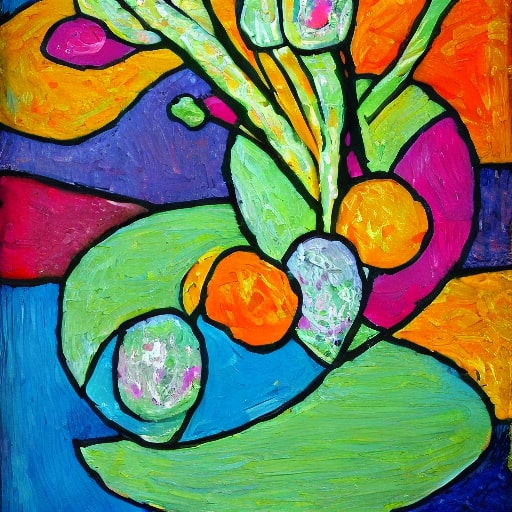} & \includegraphics[width=\linewidth,height=\linewidth]{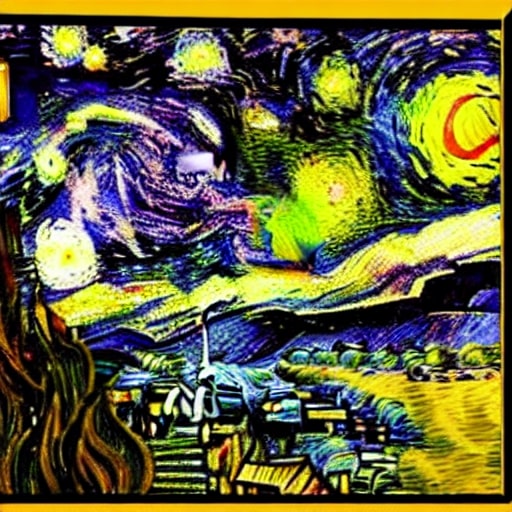} & \includegraphics[width=\linewidth,height=\linewidth]{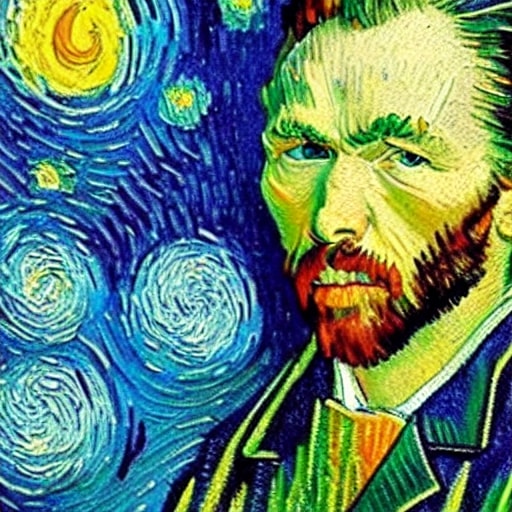} & \includegraphics[width=\linewidth,height=\linewidth]{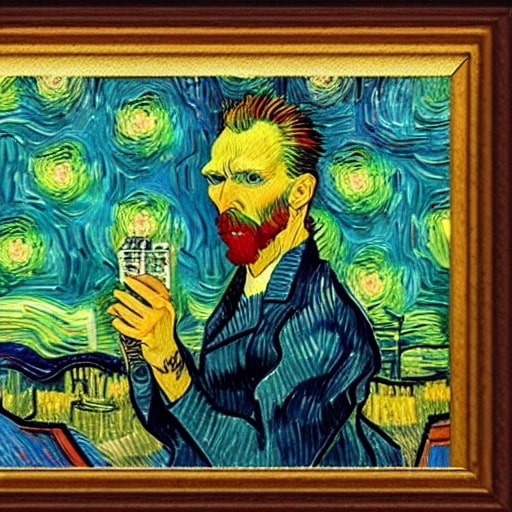} & \includegraphics[width=\linewidth,height=\linewidth]{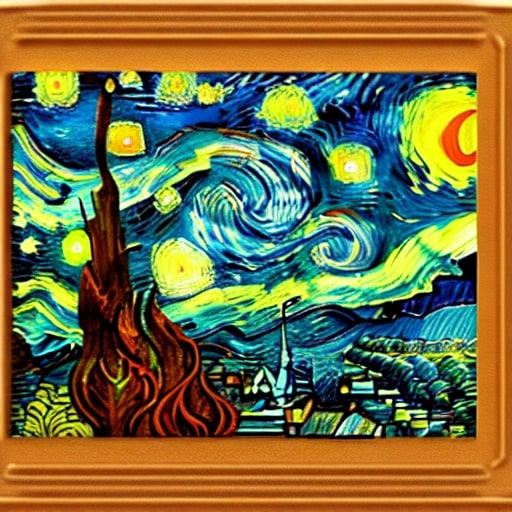} & 
\includegraphics[width=\linewidth,height=\linewidth]{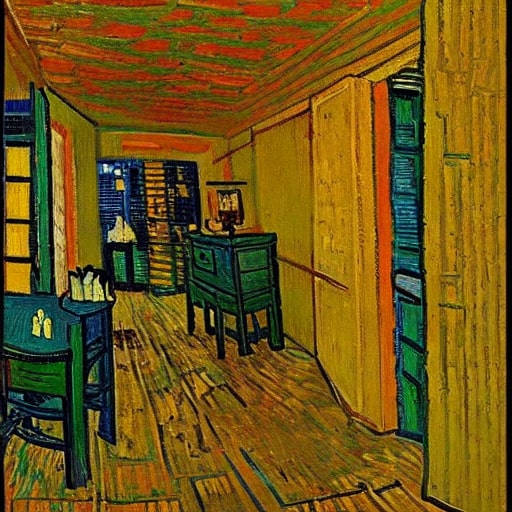} & \includegraphics[width=\linewidth,height=\linewidth]{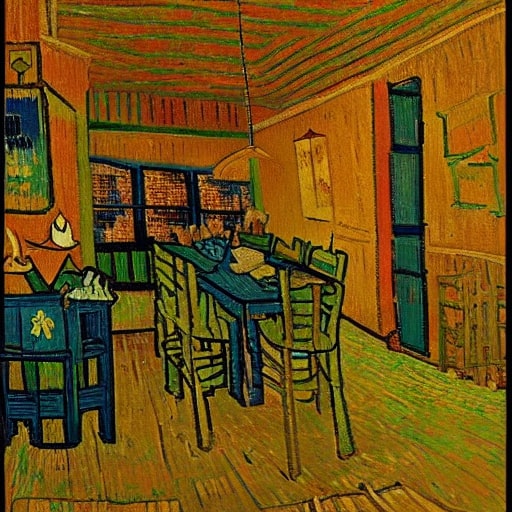} & \includegraphics[width=\linewidth,height=\linewidth]{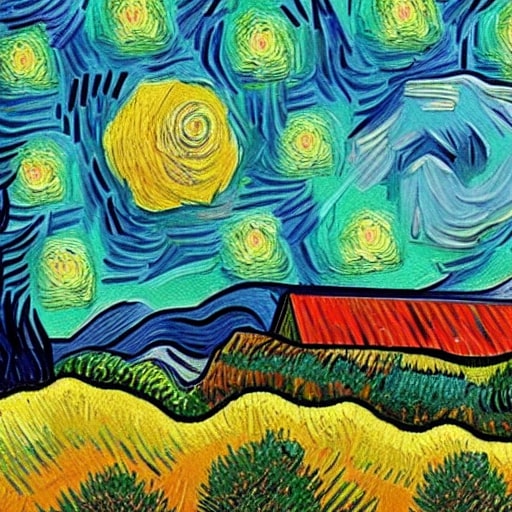} & \includegraphics[width=\linewidth,height=\linewidth]{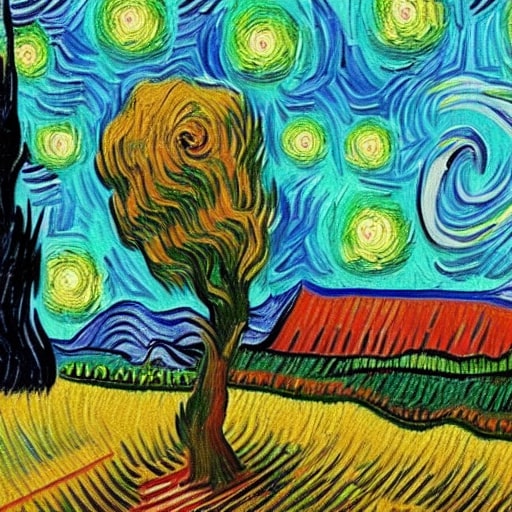} & \includegraphics[width=\linewidth,height=\linewidth]{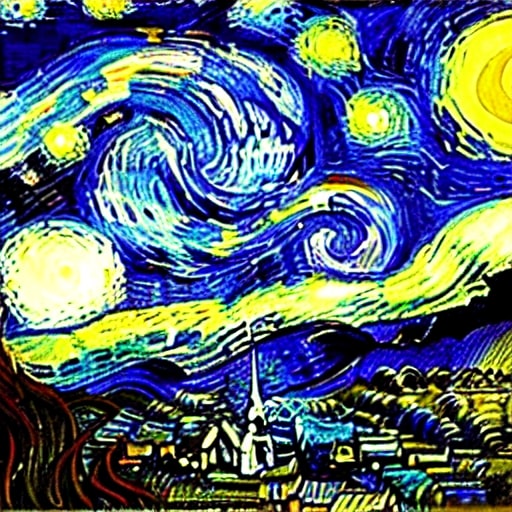} & \includegraphics[width=\linewidth,height=\linewidth]{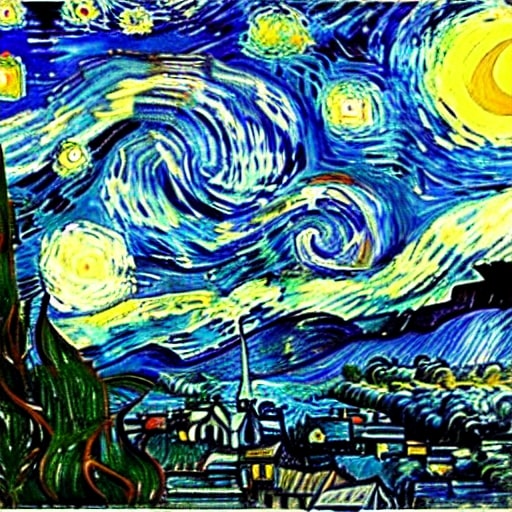} 
\end{tblr}
}
\caption{\sl\textbf{Concept erasure methods often filter out only a tiny volume in input space}. Top row: Erased Stable Diffusion (with the ``Van Gogh'' concept erased); bottom row: SD 1.4. We plot generations using various adversarially optimized prompt embeddings, located at different Cosine similarities from the embedding of the prompt ``Van Gogh''. Values in square brackets represent cosine similarities in embedding space with the prompt ``Van Gogh'' and are ordered from \textit{left} (input is far away from the concept name) to \textit{right} (closer to the concept name).  ESD continues to produce ``Van Gogh'' concepts when the input prompt is far away from the original concept name.}
\label{fig:distance_ci}
\end{figure*}

\section{Introduction}

\label{sec:intro}
The capacity of text-to-image (T2I) generative models to produce high-quality images has improved significantly over time. Consequently, growing concerns surround their potential for generating undesirable content. Such concerns include: ability to ``deepfake'' images of real people; ability to synthesize  copyrighted materials; and production of Not-Safe-For-Work (NSFW) content. A direct approach to mitigate these would be to perform data filtering, i.e., removing all images depicting undesired concepts from the model's training set. However, automatically web-scraped, massive datasets are extremely hard to filter, and imperfect filtering often compromises the safety or the legal compliance of the resulting generative models. Additionally, even if filtering were feasible, retraining already existing models from scratch due to changes in regulations is often impractical due to high costs. For brevity, we refer here to all kinds of undesirable generations as `unsafe'.

\begin{figure}[H]
\centering
\includegraphics[width=0.5\textwidth]{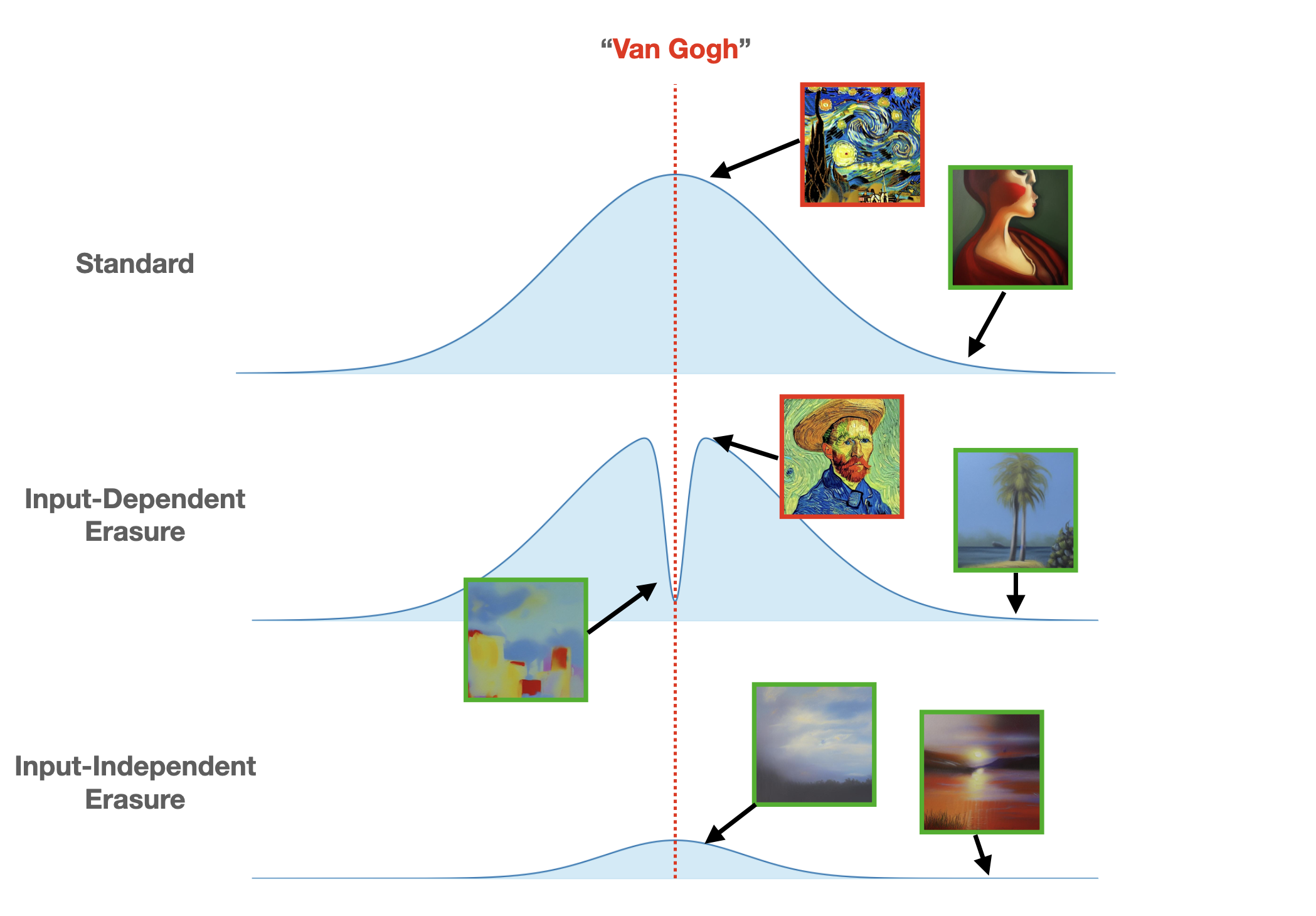}
\caption{\sl\textbf{Input-independent vs. Input-dependent concept erasure.} Illustration of the probability distribution to generate the target concept ``Van Gogh'' across the input space. Images featuring \textcolor{red}{ the ``Van Gogh''} concept are framed in red, \textcolor{green}{other images} are framed in green. Input-dependent concept erasure leaves high probability areas of generating the target concept, while input-independent erasure methods erase the target concept across the entire input space. \textbf{(Top)} In generative T2I models, the probability of generating a specific concept is high for prompt embeddings close to the concept name, but high generation probability is possible also for prompts embedding in a significant distance from it. \textbf{(Middle)} Input-dependent concept-erasure attenuates the generation probability within a small environment of the given prompt but leaves a high probability of generating the erased concept further away from the prompt embedding. \textbf{(Bottom)} Input-independent erasure attenuates the probability of generating the  target concept more consistently across the input space.}
\label{fig:your_label}
\end{figure}

Several recently proposed methods claim to ``sanitize'' unsafe concepts from T2I generative models \cite{ablating_concepts, uce, esd, negative_prompt, sld, selective_amnesia, forget_me_not}. Yet, when evaluated on unexpected inputs, these methods exhibit significant vulnerabilities, and bypassing these methods is relatively easy for adversarial methods~\cite{ring_a_bell,cce}. More specifically, most existing model sanitization approaches excel at averting the production of unsafe content, \emph{conditioned on a specific} input prompt, or sequence of tokens. However, T2I models end up learning a many-to-many mapping from prompts to image space, and a sufficiently motivated adversary --- with fairly modest compute effort --- can discover other input prompts that trigger a target unsafe concepts.

In this paper, we make further contributions towards the challenge of eliminating unsafe concepts from T2I models. Provably perfect erasure is beyond the scope of this paper, and we address a humbler challenge. Our goal is to develop an \textit{unconditional} version of  concept erasure --- namely, a method that can effectively implement concept erasure in a way that is agnostic to the choice of specific user prompts.


Our core idea is based on a recently emergent technique known as \emph{Task Vectors}~\cite{task_vector}. At a high level, a task vector (TV) represents a displacement in the model's weight space that is a result of fine-tuning; ~\cite{task_vector} shows that TVs can be flexibly used via arithmetic operations to enable editing of large models. Crucially, TV-based editing is independent of any specific user input, and therefore we showcase its ability to supply \textit{unconditional} safety to T2I models. To apply TV-based concept erasure, we first finetune the model to generate a specific concept or style, and refer to the obtained weight difference as our TV. Next, we {subtract} the TV (possibly multiplied by a scalar $\alpha$) from the original model, thereby erasing the unsafe concept.

In \cref{sec:motivation} we define a criterion we term \textit{unconditional} safety, measuring the performance of input-independent concept erasure on a model.
We hypothesize that for some models, given a long enough prompt, an effective adversarial input might always be possible  \cite{wolf2023fundamental}. As such, we limit the prompt to a maximal fixed length when measuring a method's ability to sanitize the generations. Although evaluating large models with our criterion is impractical, we can use it to show that TVs supply unconditional safety on toy models. 

Following the strong performance of TV edits for unconditional concept erasure on toy models, we investigate whether they can be applied to large T2I models without compromising the model's core functionality. Namely, we wish to apply TV edits while optimizing the trade-off between the erasure of unsafe concepts and the preservation of the model functionality. 
We characterize this trade-off by a parameter defining the edit strength, which is a scalar multiplying the vector magnitude. To tune the value of this parameter without relying on any given prompt, we propose a method called \textit{Diverse Inversion}. Diverse Inversion finds a large set of token embeddings in dense space, all aimed at generating the same concept we wish to delete. We optimize in parallel many token embeddings, each of which should induce the generation of the target concept we wish to delete in the T2I model. 

Our optimization process contains two constraints:
(i) Limiting the minimal pair-wise similarity between every pair of embeddings, to ensure diversity in the obtained set of prompts. (ii) Limiting the similarity between each embedding and the embedding of the natural language description of the concept name embedding.
Diverse Inversion allows us to tune the value of $\alpha$ in a manner that better generalizes to prompts compared to using a single input prompt.

Finally, we investigate what value of the strength parameter $\alpha$ should be used for adjusting weights in the task vector. We find that our Diverse Inversion technique allows us to find a good value of $\alpha$. Additionally, it allows us to select a subset of model weights to edit, achieving a better tradeoff between concept erasure and control task performance.

\textbf{Summary of our contributions.} \textbf{(i)} Showing that the vulnerability of current concept erasure methods is caused by their dependence on specific input prompts (\cref{sec:unconditional_safety}) \textbf{(ii)} Demonstrating TV-based editing as an efficient method for input-independent concept erasure (\cref{sec:tv_toy})  \textbf{(iii)} Proposing Diverse Inversion, an algorithm to find a diverse set of dense prompts corresponding to a target concept, and utilizing it to allow a better trade-off between concept-erasure and model performance (\cref{sec:method}).

\section{Related Work}
\textbf{Denoising Diffusion Models.}
Diffusion models are a class of generative models that iteratively refine a distribution through a Markov-based denoising process \cite{ddpm, nonequilibrium_thermodynamics}.
The process starts with a noise vector, $x_T$, and progressively denoises it over $T$ steps to reconstruct the original data $x_0$. In practice, the model is trained to predict the noise, $\epsilon_t$, at each timestep, $t$, which is used to progressively denoise the image, $x_t$. Latent diffusion models (LDM) \cite{ldm} enhance efficiency by working in a lower-dimensional space learned by an autoencoder. The first component of LDM includes a pre-trained encoder, $\mathcal{E}$, and decoder, $\mathcal{D}$, trained on a large dataset of images. The encoder maps an image, $x$, to a spatial latent code, $z = \mathcal{E}(x)$, and the decoder reconstructs the original image from the latent code, $\mathcal{D}(\mathcal{E}(x)) \approx x$. The second component is a diffusion model trained to generate codes in the learned latent space. Given an input, $c$, the LDM is trained to generate an image conditioned on $c$ using the following objective function:
\begin{equation*}
    \mathcal{L} = \mathbb{E}_{z \sim \mathcal{E}(x), t, c, \epsilon \sim \mathcal{N}(0,1)} \Big [ \| \epsilon - \epsilon_\theta(z_t,c,t) \|_2^2 \Big ]
\end{equation*}
where $z_t$ is the latent code for time $t$, and $\epsilon_\theta$ is the denoising network. During inference, a random noise tensor is sampled in latent space and gradually denoised to produce a latent code, $z'$. The latent code is then transformed into an image using the pre-trained decoder, $x' = \mathcal{D}(z')$.

\textbf{Concept-Erasure on T2I Models.} Recently, several strategies have been developed to prevent generative models from producing undesirable images. Negative Prompt (NP) \cite{negative_prompt} and Safe Latent Diffusion (SLD) \cite{sld} suggest modifying the inference process to divert the final output from undesired concepts. Other approaches employ classifiers to alter the output \cite{redteaming_sd_filter,sd2_release,nudenet}. Since inference guiding methods can be evaded with sufficient access to model parameters \cite{remove_nsfw_filter}, subsequent works including Erased Stable Diffusion (ESD) \cite{esd}, Selective Amnesia (SA) \cite{selective_amnesia}, Forget-Me-Not (FMN) \cite{forget_me_not}, Ablating Concepts (AC) \cite{ablating_concepts}, and Unified Concept Editing (UCE) \cite{uce} advocate for fine-tuning Stable Diffusion model weights. 

\textbf{Jailbraking Generative Models.} 
Deep neural networks are known for their brittleness and various algorithms are known for creating inputs that lead these models to produce undesirable outputs. In the context of Large Language Models (LLMs), the term ``jailbreaks'' refers to adversarial inputs that trigger unsafe, harmful, or unwanted responses from the model. Some jailbreaks have been discovered manually through experimentation or red-teaming \cite{jailbreak_in_20}, while others have been discovered through LLM generation \cite{redteaming_llm, jailbreaking_gpt4v}. Jailbreaking techniques include adding or prefixing \textit{adversarial strings} to the original request. In the realm of text-to-image models, despite undergoing research on concept erasure methods used to remove undesirable concepts from the weights \cite{esd, uce, ablating_concepts, negative_prompt, sld, forget_me_not, selective_amnesia}, recent works have shown that they are still susceptible to adversarial inputs \cite{ring_a_bell,cce}.

As current concept erasure methods for T2I models are often reliant on protecting against specific user inputs, adversarial methods find other inputs that can induce unsafe generations. In particular, Tsai \etal \cite{ring_a_bell} uses a CLIP text encoder to construct a concept vector; a vector in embedding space representing the unsafe content. It then uses a genetic algorithm \cite{genetic_algorithm} to find hard prompts that produce the concept vector in the embedding space.
Additionally, Pham \etal \cite{cce} propose Concept Inversion, which is a method based on Textual Inversion \cite{textual_inversion} to search for word embeddings that circumvent concept erasure methods. Textual Inversion \cite{textual_inversion} learns to capture the user-provided concept by representing it through new ``words'' in the embedding space of a frozen T2I model without changing the model weights. In particular, the authors designate a placeholder string, $c_*$, to represent the new concept the user wishes to learn. They replace the vector associated with the tokenized string with a learned embedding $v_*$, in essence ``injecting'' the concept into the model vocabulary. The technique is referred to as Textual Inversion and consists of finding an approximate solution to the following optimization problem:
\begin{equation*}
    v_* = \argmin_v \mathbb{E}_{z\sim\mathcal{E}(x),c_*,\epsilon\sim\mathcal{N}(0,1),t} \Big [ \| \epsilon - \epsilon_\theta(z_t,c_*,t) \|_2^2 \Big ].
\end{equation*}
\textbf{Task Vectors and Parameter Space Interpolations.} Although neural networks are inherently non-linear, previous research has shown that interpolating the weights of two neural networks can preserve their high accuracy if they share a portion of their optimization trajectory \cite{avg_weight_wide_optima, lottery_ticket}. In the context of fine-tuning, accuracy consistently improves when the weights of a pre-trained model are gradually shifted towards its fine-tuned counterpart \cite{merging_fisher,patching_open_vocab,robust_finetuning}. Beyond a single task, Matena \& Raffel \cite{merging_fisher} discovered that averaging the weights of multiple models, fine-tuned on different tasks from the same starting point can result in a model with high accuracy on all the fine-tuning tasks. Li \etal \cite{branch_train_merge} observed similar outcomes when averaging the parameters of language models fine-tuned across various domains. Wortsman \etal \cite{model_soup} found that averaging the weights of models fine-tuned on multiple tasks can improve accuracy on a new downstream task without additional training. 

Interestingly, the weight difference learned during fine tuning can also be learned on one task and transferred to another to achieve a similar function. Like a vector, it can also be multiplied by a (possibly negative) scalar, and often conveys an appropriate meaning to the model function. Ilharco \etal \cite{task_vector} first compute a Task Vector (TV) as:

$$\tau = \theta_{ft} - \theta_{pre},$$
where $\theta_{pre}$ is the pre-trained model and $\theta_{ft}$ is the model fine-tuned on a selected set of tasks. Subtracting the TV, scaled by a constant $\alpha$, from the pre-trained weights $\theta_{pre}$ will make the model perform worse on the selected tasks for which the fine-tuning process was done. On the other hand, adding a scaled TV will improve the model's performance on the same tasks. Ilharco \etal \cite{task_vector} show that Task Vectors, scaled by $\alpha \in [0, 1]$, can be applied to CLIP classifiers and LLMs to alter their behavior. In this work, we show that Task Vectors can also be applied to text-to-image diffusion models (in particular, the UNet module in Stable Diffusion) to perform concept erasure. 
\section{Conditional and Unconditional Concept Erasure}
\label{sec:motivation}

\subsection{Motivating analysis}

We start by noticing that current concept erasure methods are input-dependent. Such methods rely on the concept name to suppress the generation of a targeted concept. For instance, ESD \cite{esd} fine-tunes the pre-trained diffusion U-Net model weights to remove a specific style or concept when conditioned on a specific prompt. The authors propose a fine-tuning loss that reduces the probability of generating an image $x$ based on the likelihood described by the textual description of the concept, aiming to reduce the probability of the target concept $c$: $ \mathbb{P}_{\theta^*}(x) \propto \frac{\mathbb{P}_{\theta}(x)}{\mathbb{P}_{\theta}(c|x)^\eta}$, where $\theta^*$ is the UNet weights of the diffusion model, $\theta$ is the original weights, $\eta$ is a scale power factor, and $\mathbb{P}(x)$ represents the distribution generated by the original model. Since the loss function depends on the concept name $c$, we hypothesize that ESD only suppresses the generation of the targeted concept when explicitly prompted with its textual name.

To further investigate this, we inspect the input space of the SD 1.4 model, and look for different embeddings in dense space that would generate a target concept (e.g., ``Van Gogh''). To this end, we use Textual Inversion \cite{textual_inversion} and limit it to different similarity ranges from the concept name (i.e., the embedding of the string ``Van Gogh''). As can be seen in \cref{fig:distance_ci}, for the unedited model (second row, ``SD 1.4'') a large set of embeddings in dense space ranging in different similarities from the concept name, can all generate images featuring the target concept. Too far away from the concept name generations may gradually fail to reconstruct the concept.

While it is already known that common concept erasure methods may be circumvented by prompts not seen during the erasure process \cite{ring_a_bell,cce}, we find that they often filter only a small neighborhood around the embedding used for training. For example, with the ESD concept erasure method \cite{esd}, we see similar inversion capabilities across most of the model input space. The input filtration claimed by \cite{cce} is performed only in proximity to a specific input prompt, as we illustrate in \cref{fig:your_label}. Therefore, while existing methods are effective in blocking expected prompts, they are less robust to unexpected ones. Motivated by this analysis, we turn to the main goal of this paper: First, we define a notion of safety that goes beyond a specific user prompt. Second, we suggest an effective method for concept erasure that does not depend on a specific prompt.

\subsection{Marginal, conditional, and absolute safety}
\label{sec:unconditional_safety}

A straightforward way to quantify the safety of a model against undesired generations is to estimate the marginal probability of an unsafe generation. Given a generative model $G$, a prompt $p$ drawn from a distribution $D$, and the set of unsafe generations $U$, the marginal probability $P$ for unsafe generation is given as $S_{marginal}$:
\begin{equation}
S_{marginal} = P_{p \sim D} (G(p) \in U) .
\end{equation}
We note that the marginal probability $P$ is also potentially affected by the noise distribution (e.g., the random noise which is the input for a diffusion model).
Yet, explicitly calculating this \textit{marginal safety} is not possible as the empirical distribution of test prompts may be unknown. A main reason for it is that adversaries may change the prompt distribution \textit{after} the design of our model $G$ and its safety mechanisms.

As this notion of marginal safety is impractical, it is tempting to replace it with a criterion of \textit{conditional safety}. This notion takes into account a set of a few prompts supplied by the user, $C$, known to be related to the target concept we wish to erase. Our aim then would be to reduce the maximal probability of unsafe generation with any of the prompts $p \in C$:
\begin{equation}
S_{conditional} = \max_{{p} \in C}P(G(p) \in U) .
\end{equation}
Optimizing $G$ for this safety criteria would ensure that all prompts within the set $C$ would induce an unsafe behavior with a probability $S_{conditional}$ at most. This safety criterion is often optimized by most existing concept erasure methods.
Yet, optimizing this criterion would not yield any guarantee outside the set $C$. 

Therefore, we focus on a safety criterion that is independent from any user-supplied prompts we term \textit{unconditional safety}. We suggest a safety criterion limiting the probability with which unsafe generation would occur, given a constrained input complexity (e.g., the prompt length). While input length is a good parameter for input complexity in many cases, with dense embedding we may instead limit the resolution in which the dense embedding is given. 
The resolution of a continuous input vector $v \in \mathbb{R} ^d$ is closely related to the input prompt length \cite{wolf2023fundamental}. For one example, encoding a higher resolution dense embedding corresponds to more bits of information \cite{kolmogorov}.
Specifically, we denote the input complexity (measured by resolution or length) by $D_{L}$ and use it to write the unconditional safety criteria. Namely, the unconditional safety criteria $L_{uncod}$ is the minimal complexity $L$ (e.g., prompt length) for which we have a prompt $p \in D_{L}$  that induces an unsafe generation $G(p) \in U$ with a probability of at least $\varepsilon$.
\begin{equation}
L_{uncod} =  \text{ the minimal } L \ \  \text{s.t.}: \\
 \max_{p \in D_{L}}P(G(p) \in U)
 > \varepsilon \\
 \label{eq:unconditional_safety}
\end{equation}
We discuss other possible safety criteria in \cref{sec:discussion}.

\subsection{Task Vectors for unconditional safety}
\label{sec:tv_toy}
Although calculating the unconditional safety criterion $L_{uncod}$ is impractical for large values of $L$, we can demonstrate its improvement on toy models. We hypothesize that prompt-independent concept erasure methods such as TV edits may provide better unconditional safety. To test this hypothesis we trained a toy model with a dense ``prompt'' space of dimensions $d=8$. We trained our model to generate images from the MNIST \cite{mnist} dataset (See SM for implementation details). 
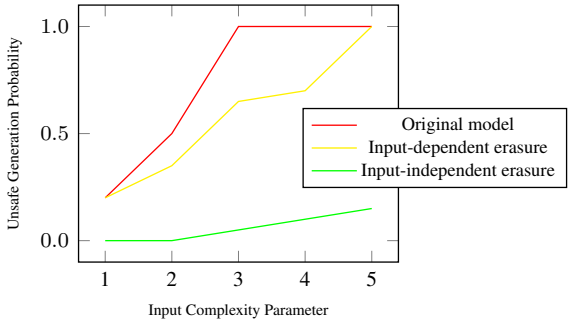
\begin{figure}[H]
    \centering
    \pgfplotsset{compat=newest}

\begin{tikzpicture}
  \begin{axis}[ 
  width=0.7\linewidth, 
  height=0.6\linewidth, 
  line width=0.5,
  grid=major,
  tick label style={font=\fontsize{8}{6}\selectfont},
  legend style={nodes={scale=0.7, transform shape}, at={(0.7,0.6)}, anchor=north west},
  label style={font=\fontsize{6}{6}\selectfont},
  grid style={white},
  xlabel={Input Complexity Parameter},
  ylabel={Unsafe Generation Probability},
  y tick label style={
    /pgf/number format/.cd,
    fixed,
    fixed zerofill,
    precision=1
  },
  xtick={1,2,3,4,5},
  xticklabels={1,2,3,4,5},
  ]
  
    \addplot[color=red] coordinates
      {(1, 0.2) (2, 0.5) (3, 1) (4, 1) (5, 1)};
      \addlegendentry{Original model}
    \addplot[color=yellow] coordinates
      {(1, 0.2) (2, 0.35) (3, 0.65) (4, 0.7) (5, 1)};
     \addlegendentry{Input-dependent erasure} 

    \addplot[green] coordinates
      {(1, 0) (2, 0) (3, 0.05) (4, 0.1) (5, 0.15)};
     \addlegendentry{Input-independent erasure}

  \end{axis}
\end{tikzpicture}
\caption{\sl\textbf{TV-based concept-erasure provides better unconditional safety.} We plot the probability of unsafe generation with the most successful adversarial prompt from each given input complexity class (See Sec.\ref{sec:unconditional_safety}). While the input-dependent (finetune-based) concept erasure method is focused on protecting against undesired generations with a specific prompt, other prompts still produce unsafe generations with high probability. The input-independent (TV-based) erasure reduces the probability of unsafe generations compared both to the original and the input-independent models, across the different complexity classes.}
\label{fig:kolmogorov}
\end{figure}

We apply three different concept-erasure models to erase the MNIST digit 0 (``the target concept'') from our diffusion model:
(i) \textit{Input dependent concept-erasure}: We fine-tune the model to produce the remaining 9 digits when given 0 as conditional input. (ii) \textit{Input-independent concept-erasure:} We utilize a TV edit for input-independent concept-erasure \cite{task_vector}. We fine-tune our model to generate only the target concept (digit 0), and then subtract the model weight change achieved by the fine-tuning process from the original model. This process is input-independent as we perform unconditional fine-tuning discarding the usage of conditional input embedding (iii) \textit{Original model:} We also evaluate the original model, without concept erasure. 
For all models, we use a pre-trained classifier to automatically evaluate whether the target concept was indeed generated. Implementation details for the classifier and all three methods, along with examples of the faithfulness of our classifier to human semantics can be found in the SM. 

We now turn to evaluate the unconditional safety criterion for the three models above. Specifically, we define the input complexity classes $D_L$ as the resolutions in which we perform an exhaustive search of the possible prompts in continuous space. For each complexity class $D_{L}$, we explore a grid in $d$ dimensions, with the inspected values in each dimension comprised of $L$ equally spaced values, totaling $L^d$ points per grid. For example, the grid point $(0.0,1.0,1.0,0.0,-1.0,...)$ belongs to a low input complexity class, while the grid point $(0.4,0.8,0.6,-0.2,-1.0,...)$ belongs to a higher one. Intuitively, the higher the input complexity class we examine, the closer we are to an exhaustive search in continuous input space.

We can clearly see in \cref{fig:kolmogorov} that the TV edit provides a much better unconditional safety $L_{uncod}$  guarantee. The original and fine-tuned models provide a non-trivial probability of generating the target concept, even for relatively low complexity parameters $L$. Moreover, even for the fine-tuned concept erasure, we can find prompts in the medium complexity range ($L=5$) that generate the target concept with very high probability. However, with the TV-based concept-erasure, the unsafe generation can be better mitigated across all examined complexity classes. This suggests that the TV-based erasure does not merely input-filter the model, but attenuates its ability to generate the unsafe concept more robustly across the input space. 

Interpreting this result according to our unconditional safety criteria (\cref{eq:unconditional_safety}), we find that only the input independent method provides a non-trivial bound of the unsafe generation probability for complexity parameters $L \geq 5$. As our input space in this experiment is of dimension $d=8$, complexity classes of large values of $L$ are infeasible to compute ($D_{L=6}$ already contains $L^d=1 679 616$ possible dense prompts).
\section{Diverse Inversion for Robust Concept Erasure Using  Task Vectors}
\label{sec:method}
Motivated by the potential of TV-based editing as a method capable of improving the unconditional safety of T2I models, we now focus on applying this technique to larger models. Namely, we wish to erase unsafe concepts from large diffusion models while otherwise retaining their text-to-image capabilities. Measuring the degree of preservation of the desired text-to-image capabilities can be done directly, since this typically involves expected user inputs and outputs. However, anticipating the model's reaction to adversarial prompts \textit{unknown} at the time of editing can be challenging. 

To estimate how well the model is protected against unexpected inputs, we would like to observe its outputs for a diverse array of adversarial prompts. We cannot inspect all the input prompts of a given length as we did for the toy model, due to the very large number of possible prompts. To this end, we create a diverse safety validation set composed of diverse input tokens that can all generate unsafe content with the original model. 
We note that a real-life adversary chooses their prompt after TV-based  concept erasure has been applied, and not before it. Yet, the fact that an adversarial prompt often transfers well between erased and original (un-erased) models \cite{cce,gcg} motivates us to rely on a large set of diverse adversarial prompts optimized for the original method.  We show in Sec.\ref{sec:results} that our method allows us to apply TV-based erasure to gain robustness to adversarial techniques applied after the erasure edit. 

Our method for applying Task Vectors for concept erasure in large models consists of three parts. First, we learn a diverse set of adversarial prompts, allowing us to estimate TV edit robustness. Next, we show this learned set allows us to choose robust hyper-parameters for TV edits while maintaining the model utility. Finally, we show that we can not only choose performant hyper-parameter values but also sub-select the set of model parameters we wish to edit for better performance.

\subsection{Diverse Inversion}
As we discuss in \cref{sec:motivation}, concept erasure methods can provide a false sense of security by performing ``input-filtering''. This suggests that additional inputs are needed to better evaluate concept erasure methods.
We would like to have a diverse set on inputs, evaluating the concept erasure capability independently from any specific adversarial prompt. 
Yet, our experiment in Fig. \ref{fig:distance_ci} also shows that in addition to suppressing the harmful generation when prompted with the concept name, the sanitized Stable Diffusion model also sanitizes surrounding word embeddings. Therefore, these additional inputs need to be far from the embedding of the concept name as well as sufficiently diverse. To create a better list of inputs for robust evaluation of concept erasure, one can search for word embeddings as follows:
\begin{equation}
    \begin{aligned}
        &\mathbf{v}_* = \argmin_{\mathbf{v}} \mathbb{E}_{z\sim\mathcal{E}(x),c_*,\epsilon\sim\mathcal{N}(0,1),t} \Big [ \| \epsilon - \epsilon_\theta(z_t,c_*,t) \|_2^2 \Big ], \\
        &\text{s.t. } \left\{
        \begin{aligned}
            &\text{Sim}(v_{i*}, v_{\text{concept}}) \in [a,b] \text{for} i = 1, 2, \ldots, n, \\
            &\text{Sim}(v_{i*}, v_{j*}) \in [c,d] \text{for} i, j = 1, 2, \ldots, n, i \neq j .
        \end{aligned}
        \right.
    \end{aligned}
    \label{eq:multi_concept_inversion}
\end{equation}
In Eq. \ref{eq:multi_concept_inversion}, we optimize for a set of embeddings $\mathbf{v}_* = (v_1, v_2, ..., v_n)$. The first constraint ensures that the learned embeddings are not too close to the embedding of the concept name (e.g. Van Gogh). The second constraint pushes the learned embeddings away from each other to diversify them. Nevertheless, the optimization procedure in Eq. \ref{eq:multi_concept_inversion} can be highly non-convex, and we found that vanilla inversion with random restart can be used as an approximation to learn sufficiently diverse embeddings for our proposed erasure method.

\subsection{Tuning the TV edit strength}

With the augmented set of inputs that all make Stable Diffusion generate images of the target concept, we can choose the parameter $\alpha$ that controls the edit strength of the TV. We look for a value that suppresses any such generation with a prompt from our Diverse Inversion set. In a recent result, Pham \etal \cite{cce} show that a robust model might not always be usable in practice. I.e. the model outputs for non-adversarial prompts will not align with the prompts' semantics. Hence, we also measure the model performance on control tasks featuring unrelated concepts. Examining both measures we can vary the value of $\alpha$ to create a scatter plot and pick the $\alpha$ that yields the desired trade-off between robustness and usability (\cref{sec:results}). 

\subsection{Sub-selecting TV weights}
\label{sec:subselecting}
In our experiments, larger values of $\alpha$ tend to make the Stable Diffusion model more robust against inversion. However, this can also affect the model performance on unrelated tasks. Motivated by \cite{does_localization_inform, memorization_localized}, we hypothesize that not all layers on the UNet need to be edited. We test our hypothesis by not editing certain blocks of the UNet. In other words, we suggest pruning certain layers of the TV weights.
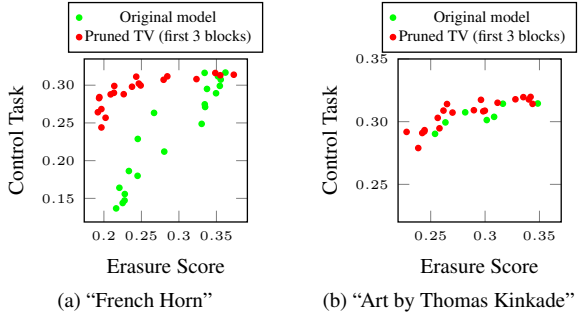
\begin{figure} [!t]
\begin{tabular}{cc}
    \begin{minipage}{0.45\linewidth}
        \centering
        \pgfplotsset{compat=newest}

\begin{tikzpicture}
  \begin{axis}[ 
  width=\linewidth, 
  height=\linewidth, 
  line width=0.5,
  grid=major,
  tick label style={font=\fontsize{5.5}{6}\selectfont},
  legend style={nodes={scale=0.6, transform shape}, at={(-0.1,1.35)}, anchor=north west},
  label style={font=\fontsize{8}{6}\selectfont},
  grid style={white},
  xlabel={Erasure Score},
  ylabel={Control Task},
  y tick label style={
    /pgf/number format/.cd,
    fixed,
    fixed zerofill,
    precision=2
  },
  ]
    \addplot[only marks, color=green, mark=*, mark size=1pt] coordinates
      {(0.33424919843673706, 0.3163394381602605) (0.3620094358921051, 0.316765539702915) (0.35200658440589905, 0.3118072492735727) (0.35576003789901733, 0.3074310758993739) (0.33742040395736694, 0.2950584065346491) (0.35453781485557556, 0.2990696838214284) (0.3494817614555359, 0.2893368463431086) (0.33493149280548096, 0.271110875975518) (0.3344448208808899, 0.2746513273034777) (0.266852468252182, 0.26329112975370317) (0.33032476902008057, 0.24874320732695715) (0.24514469504356384, 0.22878836929088547) (0.2804572582244873, 0.21200668084479513) (0.24489042162895203, 0.17989190313078107) (0.23329463601112366, 0.1862576116053831) (0.22064423561096191, 0.16405107435725985) (0.22784338891506195, 0.15575352763491018) (0.22755059599876404, 0.14722200944310143) (0.22499656677246094, 0.14385411888360977) (0.2163122445344925, 0.13672432214731262)};
      \addlegendentry{Original model}


    \addplot[only marks, color=red, mark=*, mark size=1pt] coordinates
      {(0.37314435839653015, 0.3138766835133235) (0.35515034198760986, 0.3131843950776827) (0.3484961986541748, 0.31618767728408176) (0.3234090209007263, 0.308072601045881) (0.2845328748226166, 0.3117159168635096) (0.27959105372428894, 0.3071830159141904) (0.2430916279554367, 0.31129819012823556) (0.2489987015724182, 0.2996308310400872) (0.24650996923446655, 0.30216468409413383) (0.23746660351753235, 0.29796637452784036) (0.2138412892818451, 0.2989684928740774) (0.22641335427761078, 0.2880586396370615) (0.20897802710533142, 0.28782870017346884) (0.21338164806365967, 0.28987743847426917) (0.19397416710853577, 0.28426184185913633) (0.19356173276901245, 0.28281175770929884) (0.19663643836975098, 0.26834762238320853) (0.19189828634262085, 0.2641634476326761) (0.20222243666648865, 0.25682968220540453) (0.19668836891651154, 0.24397359717459904)};
      \addlegendentry{Pruned TV (first 3 blocks)}


  \end{axis}
\end{tikzpicture}
        \subcaption{``French Horn''}
    \end{minipage} &
    \begin{minipage}{0.45\linewidth}
        \centering
        \pgfplotsset{compat=newest}

\begin{tikzpicture}
  \begin{axis}[ 
  width=\linewidth, 
  height=\linewidth, 
  line width=0.5,
  grid=major,
  tick label style={font=\fontsize{5.5}{6}\selectfont},
  legend style={nodes={scale=0.6, transform shape}, at={(-0.1,1.35)}, anchor=north west},
  label style={font=\fontsize{8}{6}\selectfont},
  grid style={white},
  xlabel={Erasure Score},
  ylabel={Control Task},
  y tick label style={
    /pgf/number format/.cd,
    fixed,
    fixed zerofill,
    precision=2
  },
  xmin=0.22, 
  xmax=0.37, 
  ymin=0.22,
  ymax=0.35,
  ]
    \addplot[only marks, color=green, mark=*, mark size=1pt] coordinates
      {(0.3484603762626648, 0.31448411799612497) (0.31636881828308105, 0.31429725956349147) (0.3083142042160034, 0.3038357978775388) (0.28172627091407776, 0.30744854502734686) (0.30153921246528625, 0.3012846378343446) (0.2635502517223358, 0.29940185092744376) (0.2539857029914856, 0.2902806954724448) (0.2182752639055252, 0.2847607313167481) (0.21158960461616516, 0.2850209886119479) (0.19975410401821136, 0.2606884278357029) (0.2067822515964508, 0.2535969519189426) (0.17289121448993683, 0.23768312874294462) (0.20928119122982025, 0.2048608674889519) (0.18839676678180695, 0.19252719836575644) (0.1986616998910904, 0.19059456867121516) (0.20297706127166748, 0.17499538138508797) (0.20636093616485596, 0.15873186698272115) (0.22030964493751526, 0.15905759004609926) (0.2094225287437439, 0.1503513613272281) (0.1893031746149063, 0.14735505907308488)};
      \addlegendentry{Original model}


    \addplot[only marks, color=red, mark=*, mark size=1pt] coordinates
      {(0.34183269739151, 0.3197462537458965) (0.34020674228668213, 0.317730748582454) (0.3351849317550659, 0.3195468652106467) (0.34365758299827576, 0.31417800876356305) (0.32790106534957886, 0.317833715961093) (0.3114638030529022, 0.3150717942487626) (0.2962314486503601, 0.31740827716532205) (0.29978951811790466, 0.30875272268340703) (0.28970178961753845, 0.30911032465242205) (0.2983032763004303, 0.30824855111894156) (0.2700521945953369, 0.30724718279781793) (0.26174163818359375, 0.3088305989901225) (0.265006959438324, 0.3140623306944257) (0.2564553916454315, 0.3030197960989816) (0.24435831606388092, 0.29322566446803866) (0.2580668330192566, 0.2946926039599237) (0.2441219687461853, 0.29218651567186626) (0.2422339916229248, 0.2909472393138068) (0.2278357595205307, 0.2918252313420886) (0.23855657875537872, 0.2789580939071519)};
      \addlegendentry{Pruned TV (first 3 blocks)}

  \end{axis}
\end{tikzpicture}
        \subcaption{``Art by Thomas Kinkade''}
    \end{minipage} 
\end{tabular}
\caption{\sl\textbf{The trade-off between erasure score and control task performance.} We plot the robustness measured according to erasure score, \textbf{lower is better}, and control task performance, \textbf{higher is better}, for models erased with different TV edit strengths (parameterized by $\alpha$).
Our Diverse Inversion method allows us to explore the trade-off between concept erasure robustness and model utility when editing different subsets of the model parameter. 
We discover that different target concepts may benefit from editing different subsets of model parameters.}
\label{fig:ti_prune}
\end{figure}
\section{Experiments}
\subsection{Experimental setup}
\textit{Metrics:} To assess the content of the generated images, we use CLIP ViT-B/32 \cite{clip} pre-trained on LAION-2B \cite{laion5b}. Following previous works \cite{cce,selective_amnesia,esd,uce}, the control task for all experiments is the average CLIP similarity score of 6 concepts across 3 different concept categories (artistic style, objects, and specific people): ``art by Kilian Eng'', ``art by Picasso'', ``garbage truck'', ``chain saw'', ``Brad Pitt'', and ``Angelina Jolie''. Motivated by our experiment in Sec. \ref{sec:motivation}, we propose to use a metric known as \emph{Erasure Score} to validate the robustness of the edited Stable Diffusion model to many different attack prompts. The metric is defined as follows: after obtaining word embeddings via Diverse Inversion, we generate an image for each learned embedding and the concept name from the Stable Diffusion model. Erasure Score (ES) is defined as the maximum (calculated over all generated images) CLIP similarity between the generated images and the concept name. A lower Erasure Score indicates more robustness against adversarial inputs. Our results on robustness to different adversarial methods are demonstrated qualitatively in \cref{fig:ring_a_bell_van_gogh,fig:ti_tv}, and quantitatively in the supplementary material (SM).

\textit{Implementation Details:} For calculating the Erasure Score we use our Diverse Inversion method to find $100$ word embeddings that trigger the generation of the erased concept. These embeddings lie evenly across 5 non-overlapping intervals, where each interval represents the allowed similarity between the learned embeddings and the embedding of the concept name. Moreover, we ensure that the pairwise similarity between learned embeddings within each interval cannot be too large. To obtain the TV for concept erasure, we first fine-tune the UNet of the Stable Diffusion model on a combination of synthetic and real images of the targeted concept, using the empty string as the caption. The fine-tuned UNet is then used to compute TV for the editing procedure. Further details appear in the SM. 

\subsection{Results}
\label{sec:results}
\begin{figure*}[!t]
\centering
\fontsize{6}{6}\selectfont
\begin{minipage}{0.35\textwidth}
\resizebox{\textwidth}{!}{
\begin{tblr}{
  width = \linewidth,
  colspec = {Q[20]Q[70]Q[70]Q[70]}, 
  column{even} = {c},
  column{odd} = {c},
  rowsep=1pt,
  colsep=1pt,
}
& ``Van Gogh'' & Concept Inversion & ``Thomas Kinkade''    \\
\begin{sideways}\hspace{10pt}\makecell{Original Model}\end{sideways} & \includegraphics[width=\linewidth,height=\linewidth]{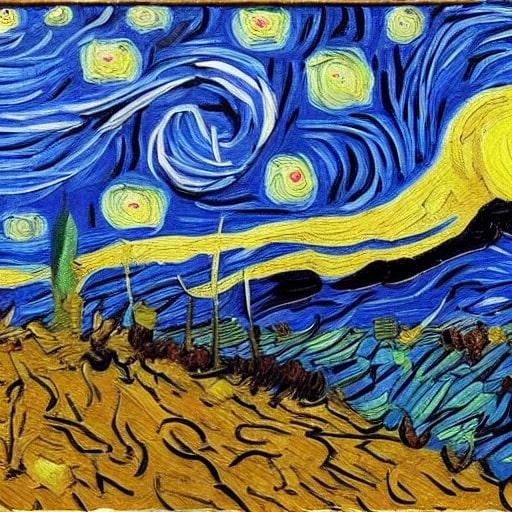} & \includegraphics[width=\linewidth,height=\linewidth]{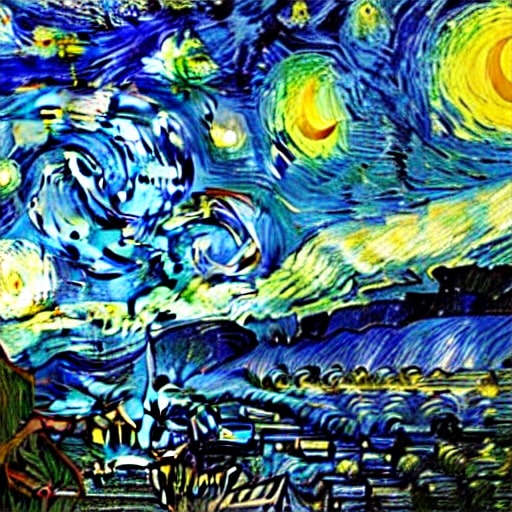} & \includegraphics[width=\linewidth,height=\linewidth]{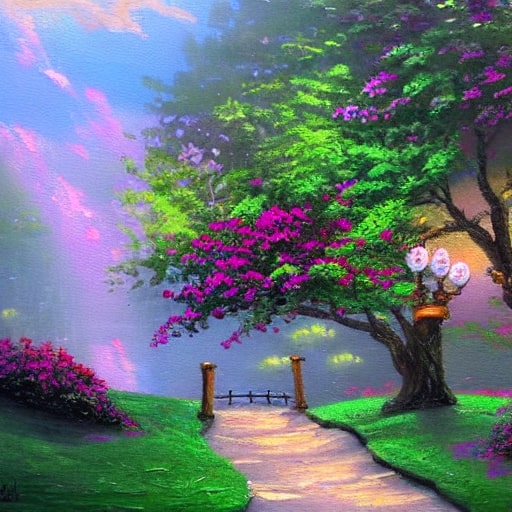} \\
\begin{sideways}\hspace{7pt}\makecell{$\alpha = 0.75$ \\ ES = $0.3409$}\end{sideways} & \includegraphics[width=\linewidth,height=\linewidth]{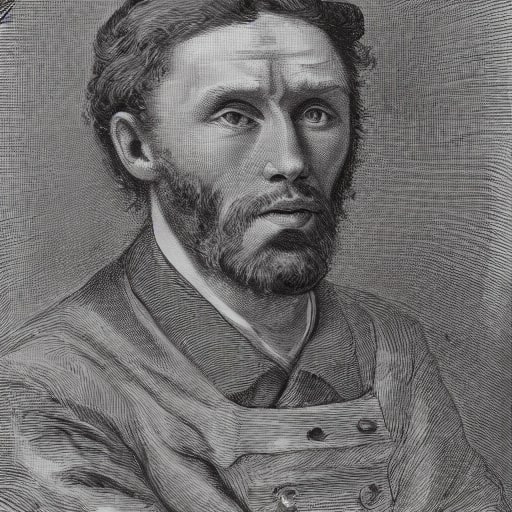} & \includegraphics[width=\linewidth,height=\linewidth]{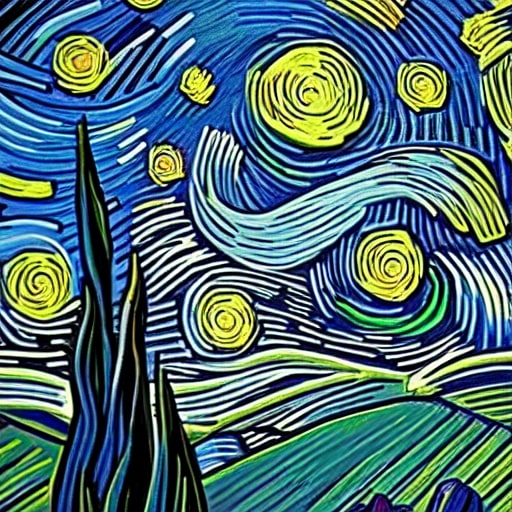} & \includegraphics[width=\linewidth,height=\linewidth]{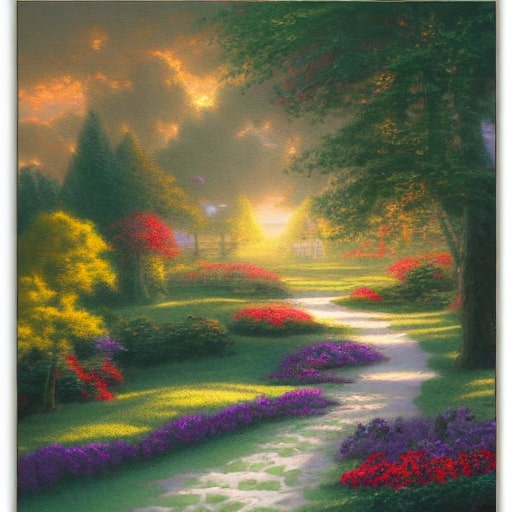} \\
\begin{sideways}\hspace{7pt}\makecell{$\alpha = 2.25$ \\ ES = $0.2376$}\end{sideways} & \includegraphics[width=\linewidth,height=\linewidth]{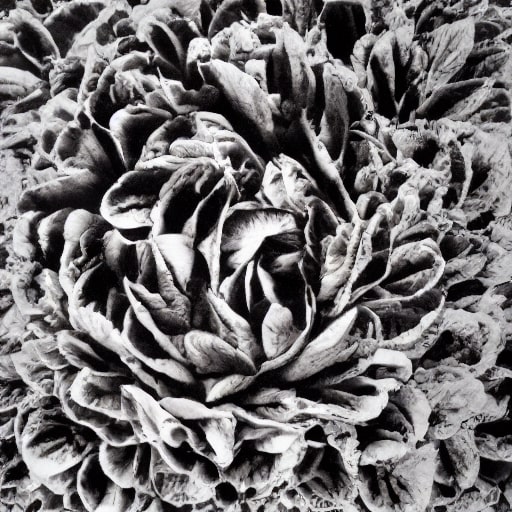} & \includegraphics[width=\linewidth,height=\linewidth]{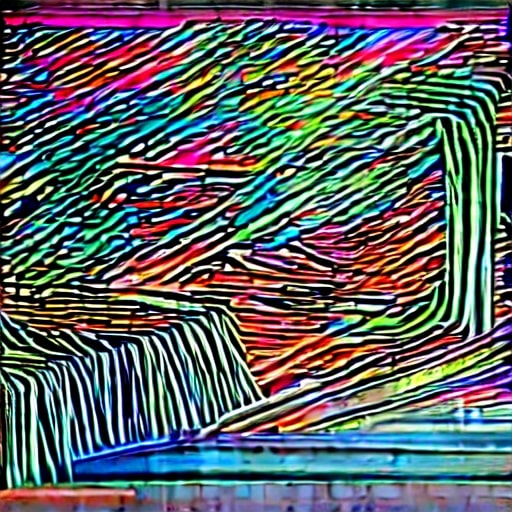} & \includegraphics[width=\linewidth,height=\linewidth]{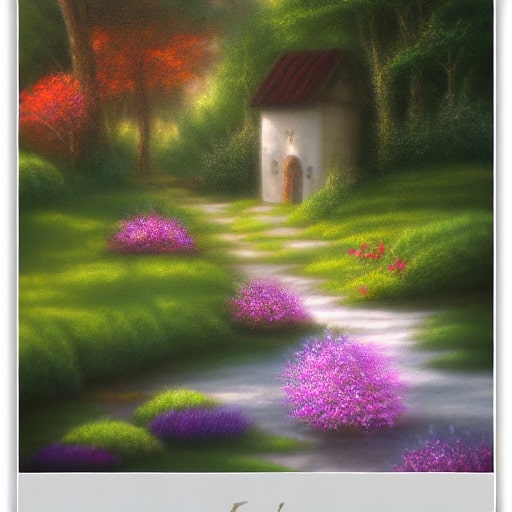}
\end{tblr}
}
\end{minipage}
\hfill
\begin{minipage}{0.35\textwidth}
\resizebox{\textwidth}{!}{
\begin{tblr}{
  width = \linewidth,
  colspec = {Q[20]Q[70]Q[70]Q[70]}, 
  column{even} = {c},
  column{odd} = {c},
  rowsep=1pt,
  colsep=1pt,
}
& ``French Horn'' & Concept Inversion & ``Garbage Truck''    \\
\begin{sideways}\hspace{10pt}\makecell{Original Model}\end{sideways} & \includegraphics[width=\linewidth,height=\linewidth]{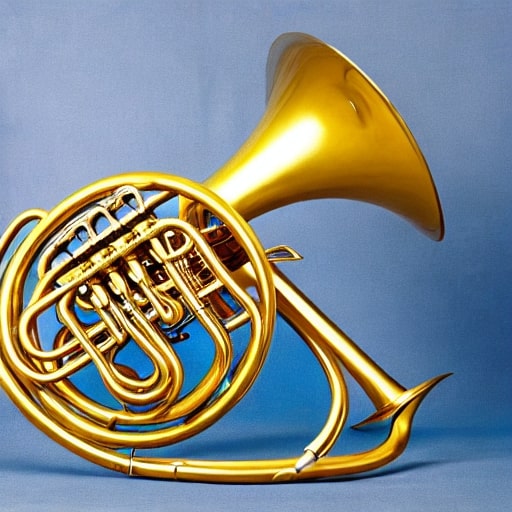} & \includegraphics[width=\linewidth,height=\linewidth]{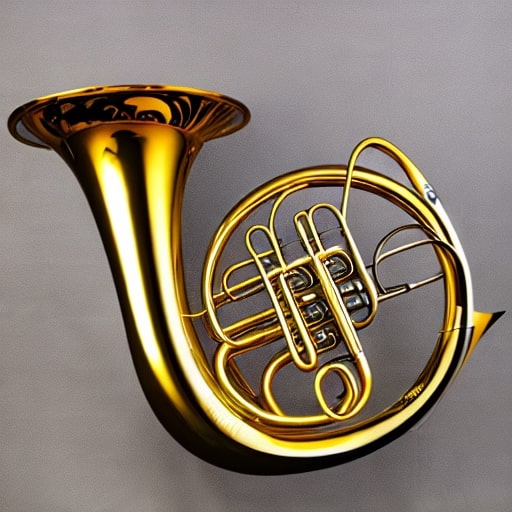} & \includegraphics[width=\linewidth,height=\linewidth]{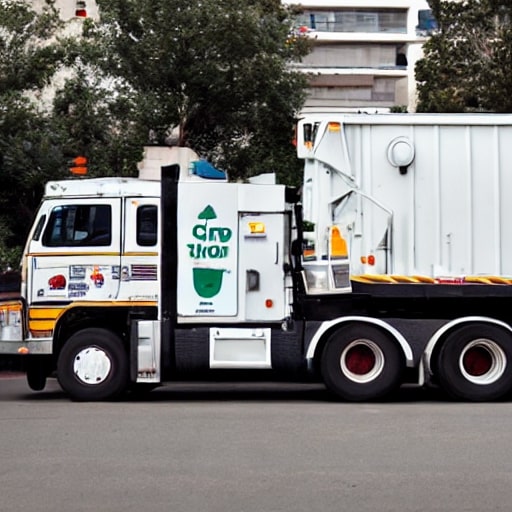} \\
\begin{sideways}\hspace{7pt}\makecell{$\alpha = 2.75$ \\ ES = $0.3344$}\end{sideways} & \includegraphics[width=\linewidth,height=\linewidth]{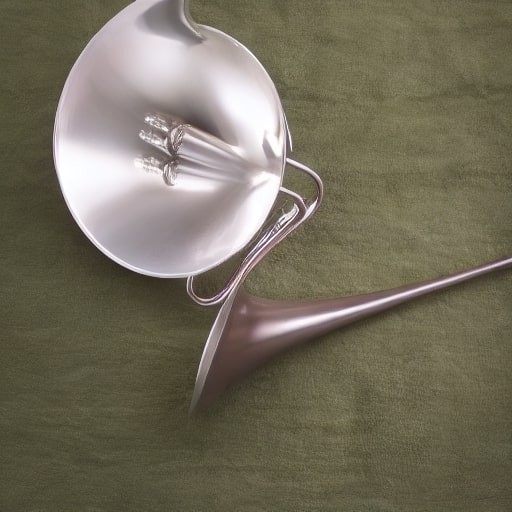} & \includegraphics[width=\linewidth,height=\linewidth]{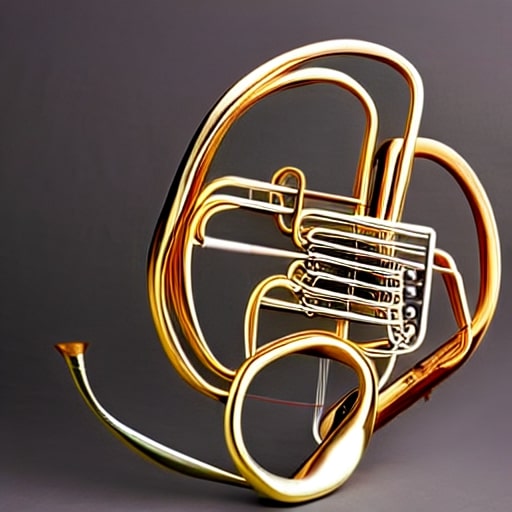} & \includegraphics[width=\linewidth,height=\linewidth]{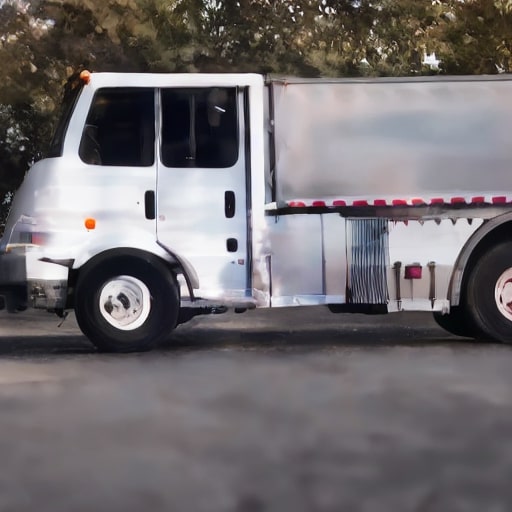} \\
\begin{sideways}\hspace{7pt}\makecell{$\alpha = 4.0$ \\ ES = $0.2344$}\end{sideways} & \includegraphics[width=\linewidth,height=\linewidth]{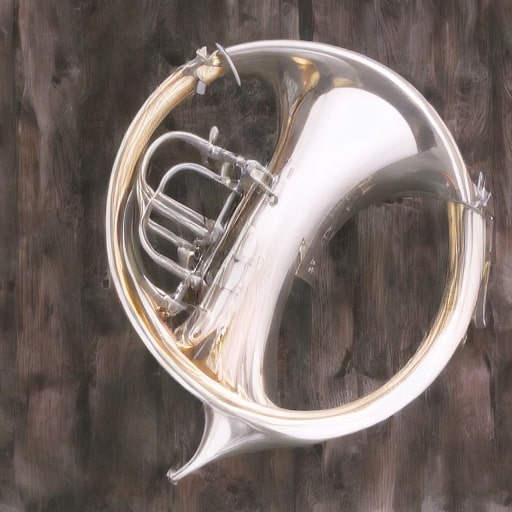} & \includegraphics[width=\linewidth,height=\linewidth]{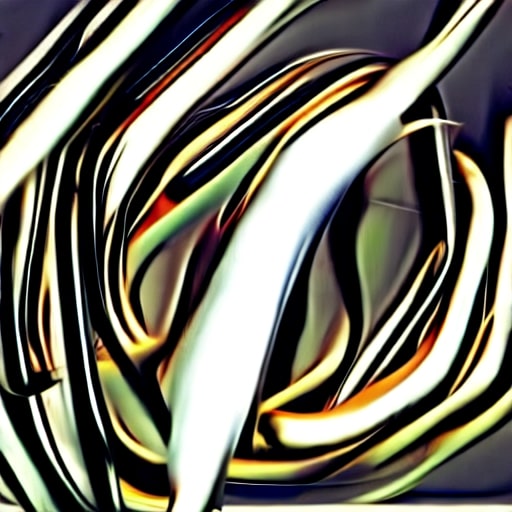} & \includegraphics[width=\linewidth,height=\linewidth]{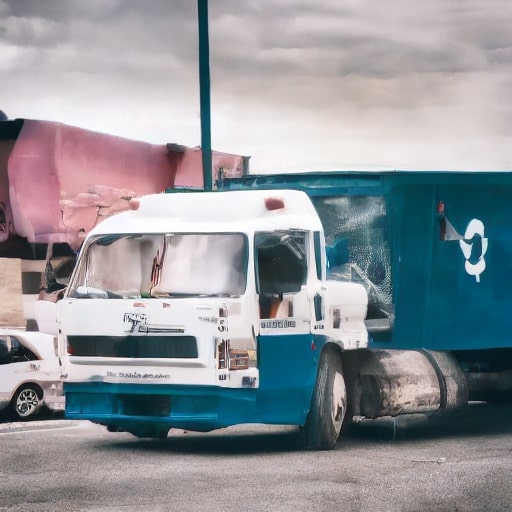}
\end{tblr}
}
\end{minipage}
\caption{\sl \textbf{TV based concept erasure robustness to Concept Inversion.} A full TV edit is utilized to erase ``Van Gogh'' \textbf{(Left)} A pruned TV  edit is utilized to erase ``French Horn'' \textbf{(Right)}. We display three model variants (by row): the original model, and two models from which we removed the targeted concept using Task Vectors of different magnitudes. In both cases, TV-based erasure is robust against Concept Inversion (\cite{cce}) and preserves the model utility on the control task. 
The third column demonstrates that TV preserves model performance on unrelated concepts.}
\label{fig:ti_tv}
\end{figure*}

\begin{figure}[!t]
  \centering
  \includegraphics[width=0.5\textwidth]{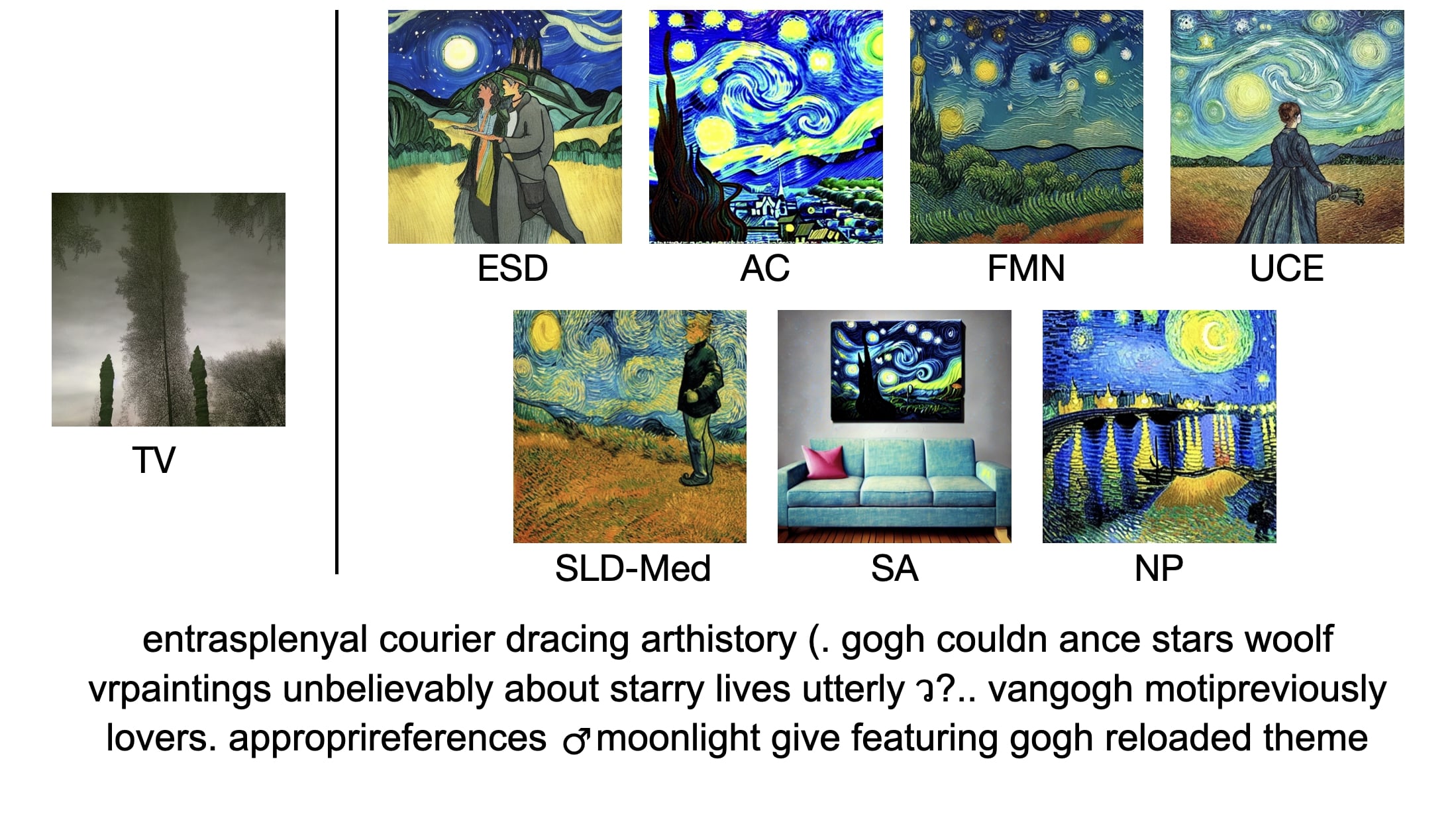}
  \caption{\sl\textbf{Generated images with the `Ring-A-Bell' \cite{ring_a_bell} prompt for the concept ``Van Gogh''.} We show that the adversarial prompt obtained from the ``Ring-A-Bell'' paper (bottom of the image) can circumvent 7 leading concept-erasure methods, but not our suggested TV erasure procedure. More similar results can be found in the SM.}
  \label{fig:ring_a_bell_van_gogh}
\end{figure}

We demonstrate in \cref{fig:ti_tv,fig:ring_a_bell_van_gogh} that our method provides robustness to current adversarial methods applied after the concept erasure edit. In the second row in each of the sub-figures of \cref{fig:ti_tv} we show that for certain values of the edit strength $\alpha$, the Stable Diffusion model manages to suppress the generation of the targeted concept when explicitly prompted with the same concept name. However, when Concept Inversion \cite{cce} is applied, we can still recover the erased concept. On the other hand, when $\alpha$ is increased, we obtain both a lower Erasure Score (ES) and a more robust erased model. This suggests that we can use the Erasure Score to guide us in selecting an appropriate edit strength, $\alpha$, to make the model more robust against adversarial inputs. We also test our edited models against hard prompts obtained from the Ring-A-Bell method \cite{ring_a_bell}. Fig. \ref{fig:ring_a_bell_van_gogh} shows that the adversarial prompts manage to fully circumvent 7 concept erasure methods but are unable to recover the target concepts erased using TV. 

A notable drawback of using Task Vectors (TV) is that this method requires using significantly higher values of $\alpha$ to enhance the Stable Diffusion model's robustness against adversarial inputs. Consequently, this might compromise the model's generative performance on concepts unrelated to the erased concept. Fig. \ref{fig:ti_prune} demonstrates that certain layers of TV can be pruned to better preserve generative performance on unrelated concepts, while still maintaining robustness against adversarial inputs.

\section{Discussion}
\label{sec:discussion}
\textbf{Model innocence as a safety goal.}
It is tempting to pursue other definitions of safety as well, such as \textit{model innocence} which stipulates that a model behaves similarly to one that was never exposed to any examples of unsafe behaviors. Yet, beyond the questions of practicality, an innocent model may still be unsafe. For example, an advanced enough model that understands the concept of something being safe-for-work, and the concept of negation, may be expected to allow the creation of various not-safe-for-work images  \cite{amodei2016concrete}.  

\textbf{Absolute safety.} Making it impossible to generate unwanted content may seem to represent the absolute ideal in model safety. However, completely avoiding unsafe content depends on the ability to recognize all unsafe behaviors. The ability to describe \textit{all} unsafe behaviors without defining such behaviors in advance is therefore left for further research \cite{amodei2016concrete}. 
\section{Limitations}
\label{sec:limitations}

\textbf{Provable guarantees for erasure.} An inherent weakness of any erasure method is the inability to evaluate them in advance against yet unknown future adversarial methods \cite{amodei2016concrete}. We acknowledge this as a weakness of our suggested method as well. Yet, we provide not only results against current adversarial methods but also a principled analysis of the unique qualities of TV-based concept erasure being input-independent; this is one of our main contributions.

\textbf{TV-based erasure.} Our suggested method is reliant on TV techniques. Yet, the parameter space of neural networks is far from being completely understood \cite{ma2020towards}. This means that the exact cases where TV-based erasure can work or fail are not clear yet. The application of Task Vectors for more fine-grained, or coarse-grained concepts, is yet to be explored. Similarly, it is not clear yet how to apply our Diverse Inversion techniques to other modalities such as language. Additionally, when erasing an excessive number of concepts it is not clear how to avoid a significant deterioration of the control task performance.

\textbf{Dependence on the Diverse Inversion set.}
While we claim to suggest an input-independent concept erasure method, our method is dependent on the discovered Diverse Inversion set. Nevertheless, we only use it to tune hyper-parameters: 
the TV edit strength, and the identity of the edited layers. Therefore, our edit is not only independent of any user-supplied prompt but also, apart from hyper-parameters, independent of the embedding found using the Diverse Inversion method.
\section{Conclusions}
We propose adapting Task Vectors (TV), a recently proposed technique for model editing, for erasing concepts from generative models. On a range of test cases, we demonstrate how TVs can be used to sanitize undesirable concepts from text-to-image models in a way that is independent of specific user prompts. This distinguishes TV from existing methods in the literature and makes it more robust. Our method, Diverse Inversion, enables us to better maintain model utility while removing harmful concepts. We anticipate that our method will be of interest to the broader AI safety community, and can be extended to other model families such as large language models (LLMs) and other multimodal vision-language models. 

\section*{Acknowledgments}
The authors were partially supported by the AI Research Institutes Program supported by NSF and USDA-NIFA under grant no. 2021-67021-35329, NSF SaTC grant 2154119, and a Cyber NYC gift from Google Research. KM was supported by a US Department of Education GAANN fellowship. NC was partially supported by the Israeli data science scholarship for outstanding postdoctoral fellows (VATAT).
\newpage
{
    \small
    \bibliographystyle{ieeenat_fullname}
    \bibliography{main}
}

\clearpage
\setcounter{page}{1}
\maketitlesupplementary

\textcolor{red}{CAUTION: This section includes generated content that may contain offensive or distressing material.}

\section{Additional Implementation Details}
For all of our experiments except the toy MNIST one, we use Stable Diffusion 1.4 (SD 1.4). To compute the TV for SD 1.4, we fine-tune the UNet component on 15 synthetic images and 15 real images obtained from Google Images (30 in total). The synthetic images are obtained from the unedited SD 1.4 using the prompt ``a photo of [\textit{object name}]'' for object concepts, and ``a painting in the style of [\textit{artist name}]'' for art style concepts. We fine-tune for 1000 steps using a learning rate of $1e-05$.

\section{Images for MNIST Experiment}
\begin{figure*}[ht]
\centering
\fontsize{8}{4}\selectfont
\resizebox{0.9\textwidth}{!}{
\begin{tblr}{
  width = \linewidth,
  colspec = {Q[40]Q[80]Q[80]Q[80]Q[80]Q[80]Q[80]Q[80]Q[80]Q[80]Q[80]}, 
  column{even} = {c},
  column{odd} = {c},
  rowsep=1pt,
  colsep=1pt,
  vline{3}={1-4}{},
}
& 0 (erased class) & 1 & 2 & 3 & 4 & 5 & 6 & 7 & 8 & 9    \\
\begin{sideways}\hspace{4pt}\makecell{Original}\end{sideways} & \includegraphics[width=\linewidth,height=\linewidth]{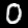} & \includegraphics[width=\linewidth,height=\linewidth]{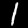} & \includegraphics[width=\linewidth,height=\linewidth]{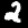} & \includegraphics[width=\linewidth,height=\linewidth]{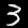} & \includegraphics[width=\linewidth,height=\linewidth]{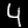}  & 
\includegraphics[width=\linewidth,height=\linewidth]{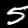} & \includegraphics[width=\linewidth,height=\linewidth]{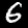} & \includegraphics[width=\linewidth,height=\linewidth]{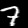} & \includegraphics[width=\linewidth,height=\linewidth]{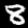} & \includegraphics[width=\linewidth,height=\linewidth]{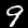} \\
\begin{sideways}\hspace{4pt}\makecell{Fine \\ tuning}\end{sideways} & \includegraphics[width=\linewidth,height=\linewidth]{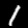} & \includegraphics[width=\linewidth,height=\linewidth]{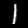} & \includegraphics[width=\linewidth,height=\linewidth]{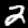} & \includegraphics[width=\linewidth,height=\linewidth]{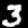} & \includegraphics[width=\linewidth,height=\linewidth]{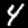} & 
\includegraphics[width=\linewidth,height=\linewidth]{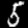} & \includegraphics[width=\linewidth,height=\linewidth]{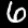} & \includegraphics[width=\linewidth,height=\linewidth]{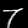} & \includegraphics[width=\linewidth,height=\linewidth]{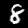} & \includegraphics[width=\linewidth,height=\linewidth]{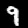} \\
\begin{sideways}\hspace{4pt}\makecell{Task \\ Vector}\end{sideways} & \includegraphics[width=\linewidth,height=\linewidth]{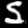} & \includegraphics[width=\linewidth,height=\linewidth]{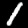} & \includegraphics[width=\linewidth,height=\linewidth]{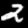} & \includegraphics[width=\linewidth,height=\linewidth]{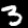} & \includegraphics[width=\linewidth,height=\linewidth]{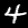} & 
\includegraphics[width=\linewidth,height=\linewidth]{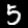} & \includegraphics[width=\linewidth,height=\linewidth]{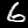} & \includegraphics[width=\linewidth,height=\linewidth]{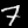} & \includegraphics[width=\linewidth,height=\linewidth]{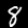} & \includegraphics[width=\linewidth,height=\linewidth]{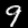}
\end{tblr}
}
\caption{\sl\textbf{ Fine-tuning and Task Vector can suppress the generation of digit 0 when the diffusion model is given class 0 as the conditional input.}}
\label{fig:toy_mnist_examples}
\end{figure*}

\begin{figure}[H]
\centering
\fontsize{20}{4}\selectfont
\resizebox{0.4\textwidth}{!}{
\begin{tblr}{
  width = \linewidth,
  colspec = {Q[80]Q[80]Q[80]}, 
  column{even} = {c},
  column{odd} = {c},
  rowsep=1pt,
  colsep=2pt,
}
Original & Fine-tuning & TV     \\
\includegraphics[width=\linewidth,height=\linewidth]{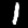} & \includegraphics[width=\linewidth,height=\linewidth]{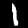} & \includegraphics[width=\linewidth,height=\linewidth]{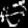}
\end{tblr}
}
\caption{\sl Inversion of the erased class (digit 0) works on the original and fine-tuned diffusion models, but not on the edited model using Task Vector.}
\label{fig:toy_mnist_ci}
\end{figure}

Fig. \ref{fig:toy_mnist_examples} demonstrates that both Fine-tuning and TV can be used to erase the digit 0 from the toy diffusion model. We also use a pre-trained classifier to quantitatively assess the generation quality of the edited models in Tab. \ref{tab:mnist_acc}. Both editing methods can erase the target class when the model is given $0$ as conditional input, while preserving generative performance on other classes. However, Fig. \ref{fig:toy_mnist_ci} shows that TV is more robust against inversion. For the toy diffusion models, we used implementation from \cite{minimal_diffusion}. However, we modify the architecture to support conditional embeddings of dimension $d = 8$, and normalize the input embeddings during training and inference. Such modifications are made to make the space of input embeddings that generate actual digits more compact. This makes the model more likely to generate faithful images when given our sampled embeddings as conditional input. We train and fine-tune using a batch size of $512$ for $100$ epochs.

\begin{table}[H]
    \centering
    \caption{\sl\textbf{Classification accuracy (\%) on generated images}}
    \begin{tabular}{|c|c|c|} \hline 
         &  Target class&  Other classes\\ \hline 
         Original&  98.2&  \textbf{98.1}\\ \hline 
 Fine-tuning& 1.4&97.3\\ \hline 
         Task Vector&  \textbf{0.4}&  97.3\\ \hline
    \end{tabular}
    
    \label{tab:mnist_acc}
\end{table}

\subsection{Ablation for Diverse Inversion}
\begin{figure}[H]
  \centering
  \includegraphics[width=0.5\textwidth]{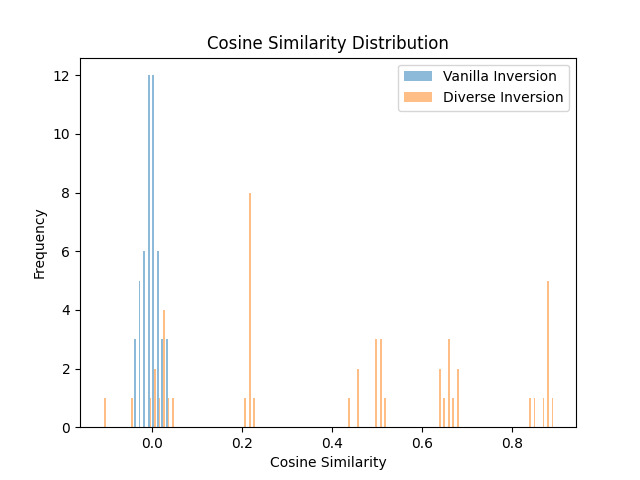}
  \caption{\sl\textbf{Histogram of cosine similarities between learned embeddings and the embedding of the concept name (``Van Gogh'').}}
  \label{fig:cosine_similariy_hist}
\end{figure}

\begin{figure*}[ht]
\centering
\fontsize{7}{6}\selectfont
\resizebox{0.9\textwidth}{!}{
\begin{tblr}{
  width = \linewidth,
  colspec = {Q[80]Q[80]Q[80]Q[80]Q[80]Q[80]Q[80]Q[80]Q[80]Q[80]}, 
  column{even} = {c},
  column{odd} = {c},
  rowsep=1pt,
  colsep=2pt,
}
Embedding 1 & Embedding 2 &  Embedding 3 & Embedding 4 & Embedding 5 & Embedding 6 & Embedding 7 & Embedding 8 & Embedding 9 & Embedding 10    \\
\includegraphics[width=\linewidth,height=\linewidth]{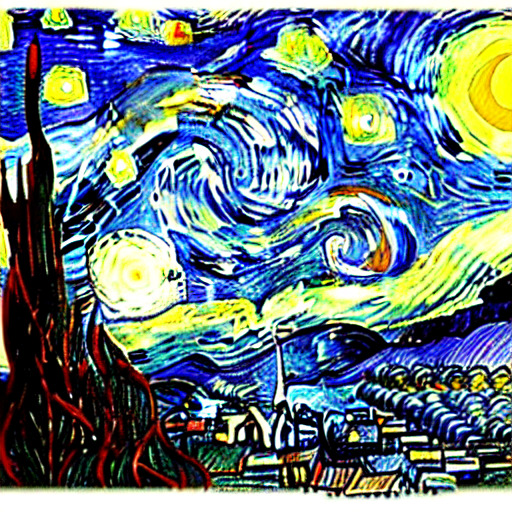} & \includegraphics[width=\linewidth,height=\linewidth]{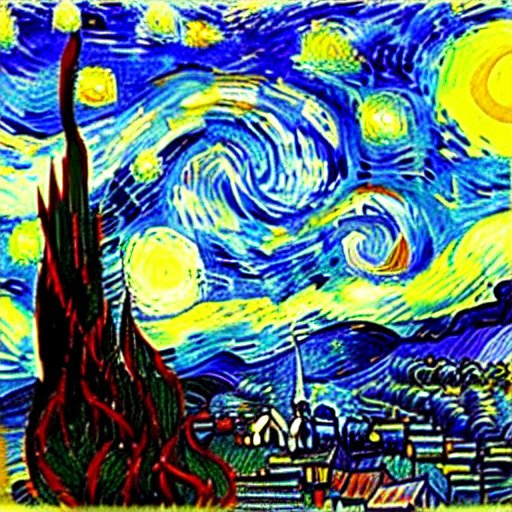} & \includegraphics[width=\linewidth,height=\linewidth]{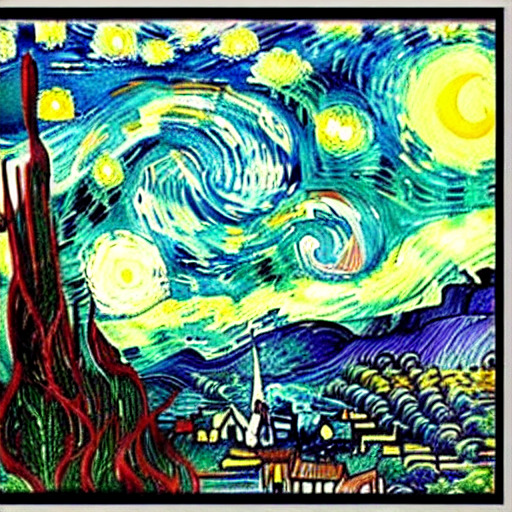} & \includegraphics[width=\linewidth,height=\linewidth]{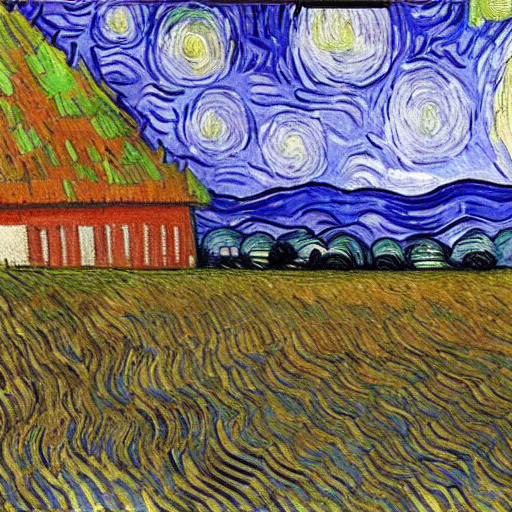} & \includegraphics[width=\linewidth,height=\linewidth]{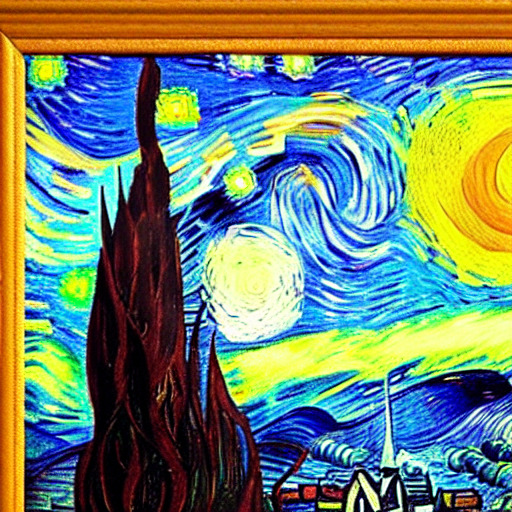}  & 
\includegraphics[width=\linewidth,height=\linewidth]{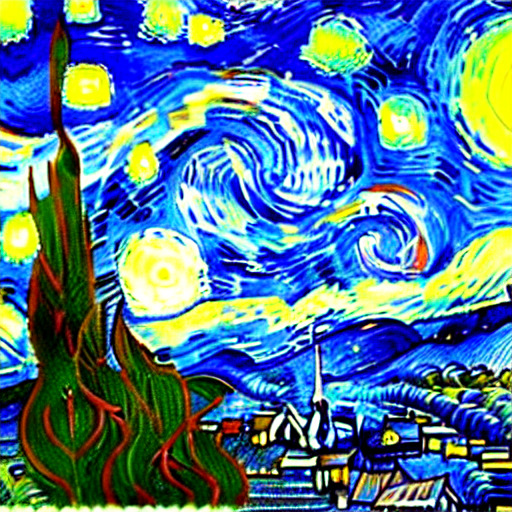} & \includegraphics[width=\linewidth,height=\linewidth]{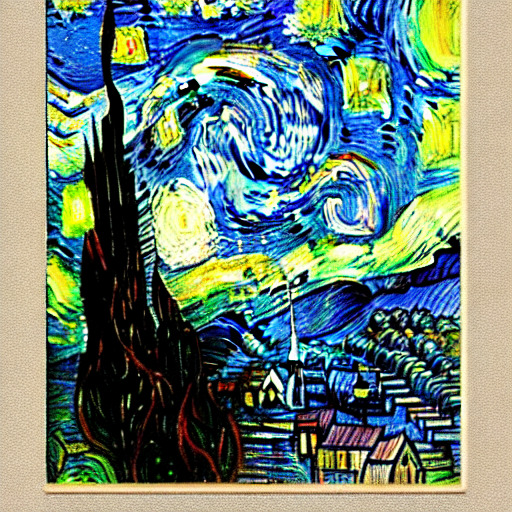} & \includegraphics[width=\linewidth,height=\linewidth]{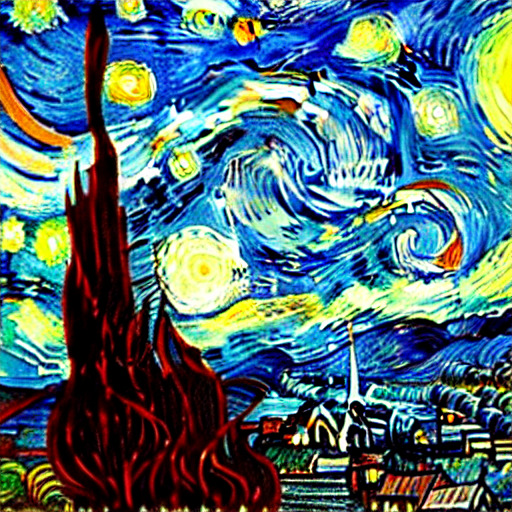} & \includegraphics[width=\linewidth,height=\linewidth]{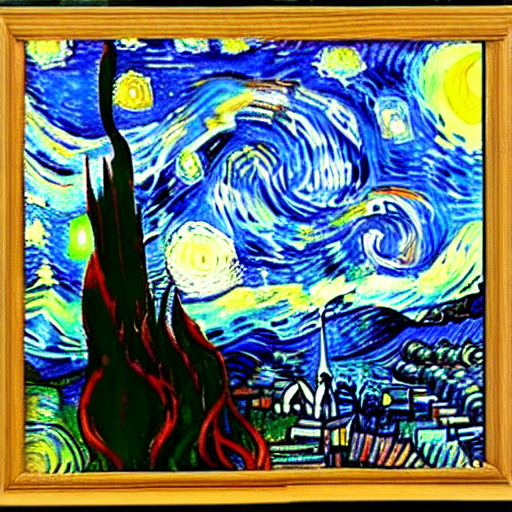} & \includegraphics[width=\linewidth,height=\linewidth]{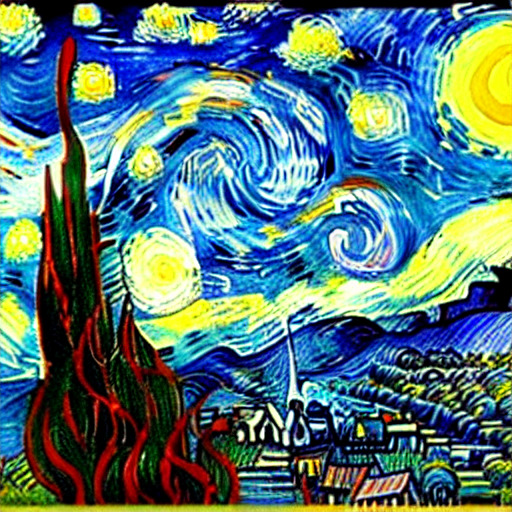}
\end{tblr}
}
\caption{\sl\textbf{Generated images using learned embeddings from Diverse Inversion}}
\label{fig:diverse_inversion_samples}
\end{figure*}

To study the necessity of Diverse Inversion, we also perform vanilla Textual Inversion (TI) \cite{textual_inversion} to find $50$ word embeddings for Van Gogh style. Fig. \ref{fig:cosine_similariy_hist} suggests that without the additional constraints, the cosine similarities between the learned embeddings through vanilla TI and the embedding of the concept name will center around $0.0$. However, with Diverse Inversion, we can enhance the diversity of our learned embeddings by controlling such cosine similarities taken with respect to the concept name. Fig. \ref{fig:diverse_inversion_samples} shows samples of SD 1.4 when we used the learned embeddings of Diverse Inversion as conditional input.


\subsection{Additional Results}
We provide additional results on the robustness of TV in Tab. \ref{tab:quantitative_object}. We followed a similar setup to Pham \etal \cite{cce} to quantify the robustness of erasure methods on objects. In particular, we first generate 500 images for the prompt with the concept name and 500 images with Concept Inversion. We then use a pre-trained classifier to measure the presence of the erased object in the generated images.

\begin{table*}[ht]
\centering
\caption{\sl\textbf{Quantitative results of Concept Inversion for object concept (Acc. \% of erased model / Acc. \% of CI)}: Using Concept Inversion, we can generate
images of the erased objects, which can be seen by an increase in average accuracy across 4 concept erasure methods except Task Vector. ``SD 1.4'' features the accuracy of the original model with the concept name. Accuracy is measured with a classifier as in \cite{esd}.}
\label{tab:quantitative_object}
\refstepcounter{table}
\begin{tblr}{
  width = \linewidth,
  colspec = {Q[150]Q[100]Q[100]Q[100]Q[100]Q[100]Q[100]},
  column{even} = {c},
  column{2-6} = {c},
  hline{1} = {2-7}{},
  hline{2} = {-}{},
  hline{12} = {-}{},
  colsep=2pt,
  rowsep=0.1pt,
}
 & SD 1.4 & TV & ESD  & UCE  & NP & SLD-Med \\
cassette player & 6.4 & 2.0 / 0.0 & 0.2 / 6.2  & 0.0 / 2.8  & 4.0 / 9.4 & 1.0 / 2.4\\
chain saw & 68.6 & 1.2 / 0.3 & 0.0 / 64.0  & 0.0 / 43.6  & 4.0 / 82.8 & 0.8 / 86.6 \\
church & 79.6 & 12.4 / 0.4 & 0.8 / 87.4  & 10.0 / 82.2  & 25.4 / 78.4 & 20.6 / 72.0 \\
english springer & 93.6 & 9.1 / 0.3 & 0.2 / 48.2  & 0.0 / 69.6  & 27.0 / 90.4 & 24.6 / 96.4 \\
french horn & 99.3 & 26.1 / 0.0 & 0.0 / 81.6  & 0.4 / 99.4 & 62.4 /  99.0 & 17.0 / 97.6 \\
garbage truck & 83.2 & 9.2 / 0.4 & 0.8 / 57.0  & 16.4 / 89.6  & 39.4 / 84.6 & 19.8 / 94.8 \\
gas pump & 76.6 & 3.2 / 0.3 & 0.0 / 73.8  & 0.0 / 73.0  & 18.0 / 79.6 & 12.8 / 75.6 \\
golf ball & 96.2 & 13.4 / 0.5 & 0.0 / 28.6  & 0.2 / 18.6  & 45.2 / 88.4 & 60.2 / 98.8\\
parachute & 96.2 & 19.2 / 0.0 & 0.0 / 94.2  & 1.6 / 94.2  & 32.8 / 77.2 & 52.8 / 95.8 \\
tench & 79.6 & 12.2 / 0.1 & 0.3 / 59.7  & 0.0 / 20.6 & 27.6 / 72.6 & 20.6 / 75.4 \\
Average  & 77.9 & 10.8 / 0.2 & 0.2 / 60.1  & 2.9 / 59.4  & 28.6 / 76.2 & 23.0 / 79.5 
\end{tblr}
\end{table*}

\begin{figure}[H]
  \centering
  \includegraphics[width=0.5\textwidth]{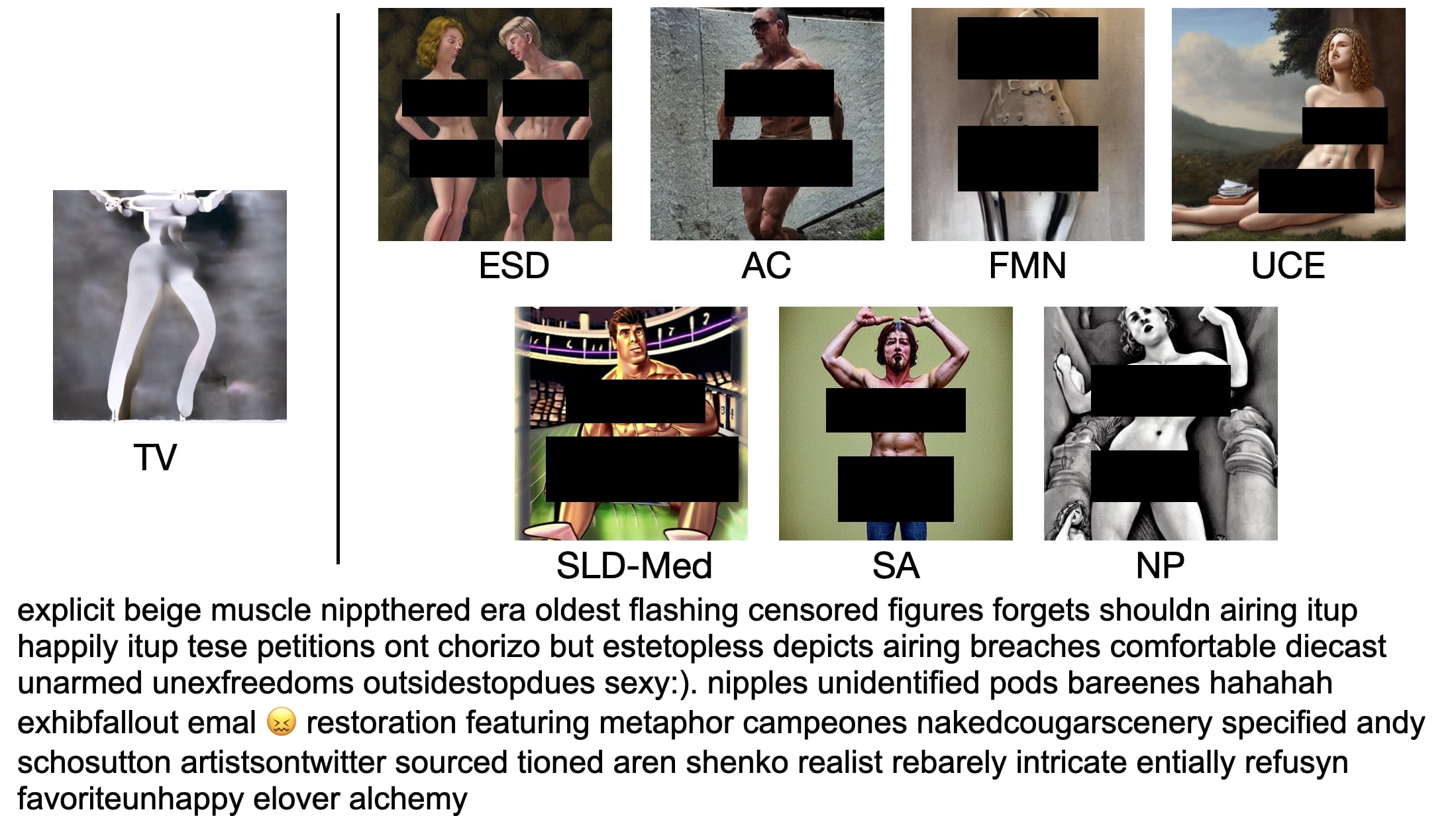}
  \caption{\sl\textbf{Generated images with the `Ring-A-Bell' \cite{ring_a_bell} prompt for the concept ``Nudity''.} We show that the adversarial prompt obtained from the ``Ring-A-Bell'' paper (bottom of the image) can circumvent 7 leading concept-erasure methods, but not our suggested TV erasure procedure.}
  \label{fig:ring_a_bell_nsfw}
\end{figure}

\begin{figure*}[ht]
\centering
\fontsize{8}{1}\selectfont
\resizebox{\textwidth}{!}{
\begin{tblr}{
  width = \linewidth,
  colspec = {Q[30]Q[80]Q[80]Q[80]Q[80]Q[80]Q[80]}, 
  column{even} = {c},
  column{odd} = {c},
  rowsep=1pt,
  colsep=1pt,
  vline{12}={1-4}{},
}
& 0 embeddings & 1 embeddings & 3 embeddings & 5 embeddings & 7 embeddings & 9 embeddings    \\
\begin{sideways}\hspace{3pt}\makecell{Concept \\ Name}\end{sideways} & \includegraphics[width=\linewidth,height=\linewidth]{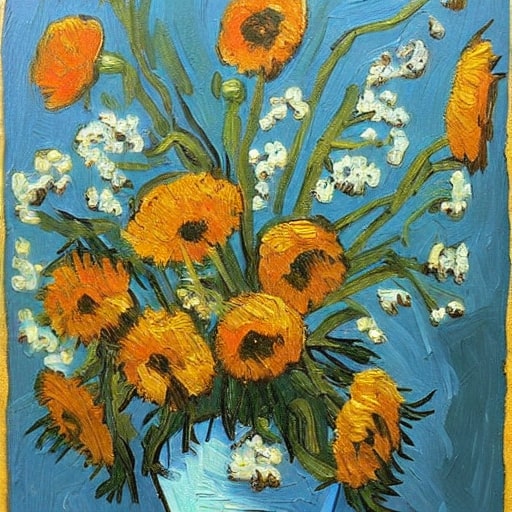} & \includegraphics[width=\linewidth,height=\linewidth]{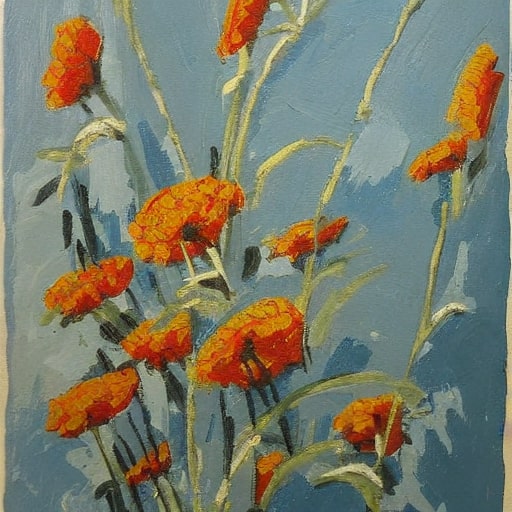} & \includegraphics[width=\linewidth,height=\linewidth]{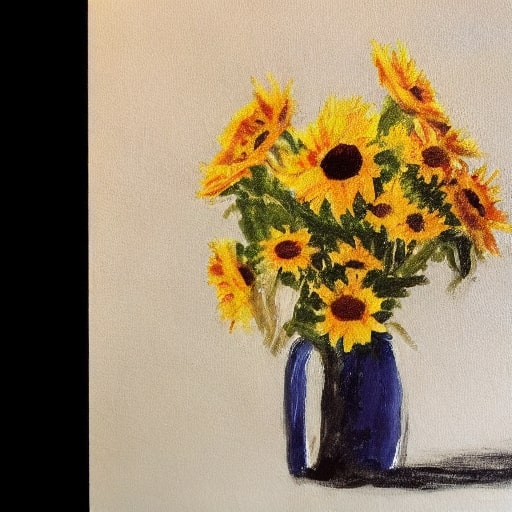} & \includegraphics[width=\linewidth,height=\linewidth]{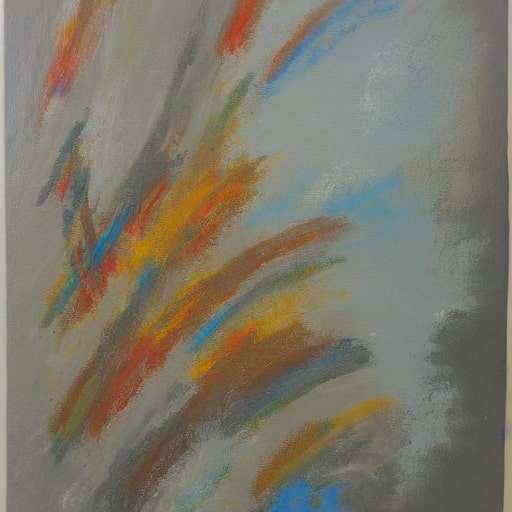} & \includegraphics[width=\linewidth,height=\linewidth]{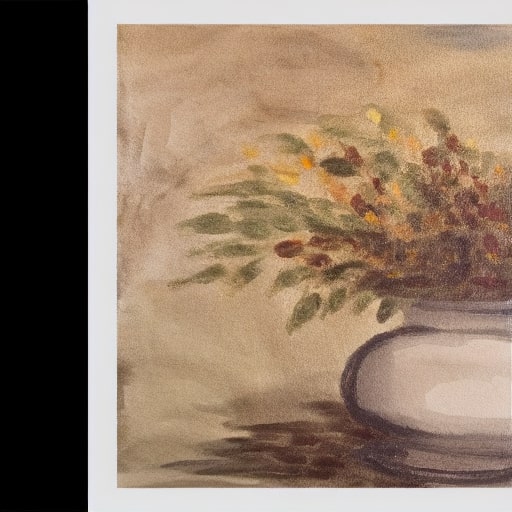} & \includegraphics[width=\linewidth,height=\linewidth]{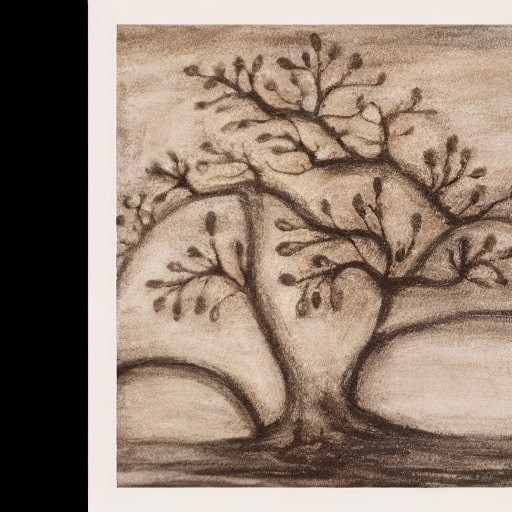} \\
\begin{sideways}\hspace{6pt}\makecell{Inversion}\end{sideways} & \includegraphics[width=\linewidth,height=\linewidth]{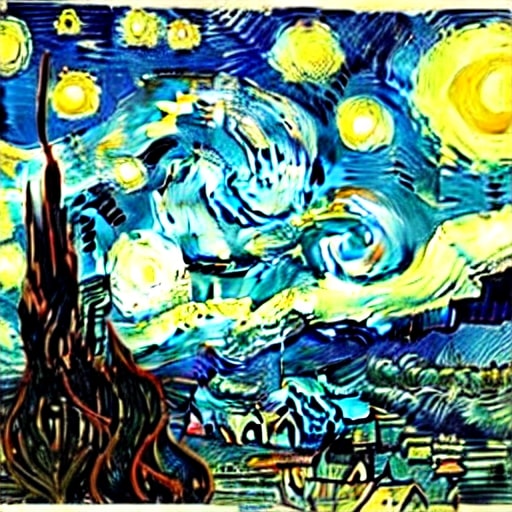} & \includegraphics[width=\linewidth,height=\linewidth]{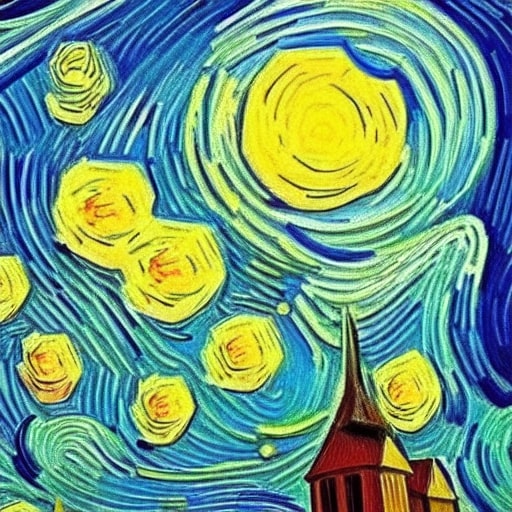} & \includegraphics[width=\linewidth,height=\linewidth]{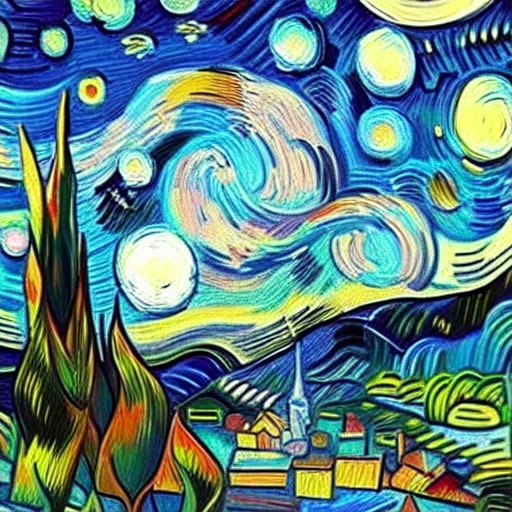} & \includegraphics[width=\linewidth,height=\linewidth]{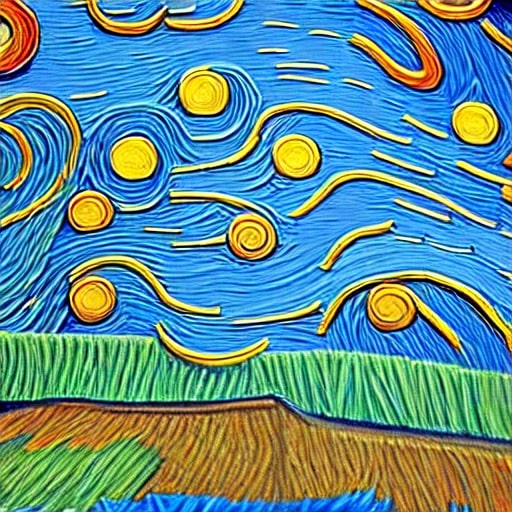} & \includegraphics[width=\linewidth,height=\linewidth]{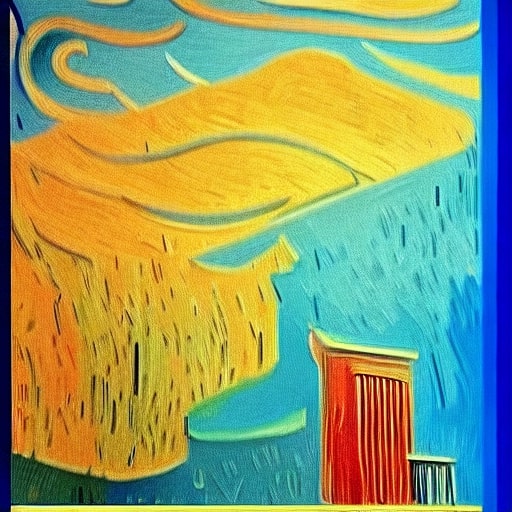} & \includegraphics[width=\linewidth,height=\linewidth]{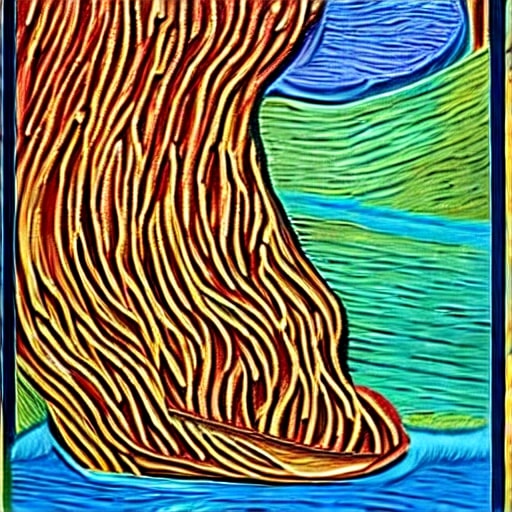}
\end{tblr}
}
\caption{\sl\textbf{More learned input embeddings can help pick a more robust $\alpha$ for TV}. Top row: Images generated using the prompt ``a painting in the style of Van Gogh''; bottom row: Images generated using the prompt ``a painting in the style of $S^{*}_{i}$'', where $S^{*}_{i}$ is the token associated with the $i^{th}$ learned embeddings. We first generate images using the concept name and learned embeddings (if any), and then choose $\alpha$ until the Erasure Score is below $0.24$. When using less than 5 learned embeddings, we can achieve a low Erasure Score even with a low $\alpha$. However, such a low ES score can provide a false sense of security since Concept Inversion can still recover the erased concept. By utilizing more embeddings, the ES score can provide a better estimate of which $\alpha$ to pick to make the model more robust against adversarial inputs.}
\label{fig:num_embeddings_ablation}
\end{figure*}

\end{document}